\documentclass{fodsOF}
\usepackage{amssymb}
  \usepackage{paralist}
  \usepackage{graphics} %% add this and next lines if pictures should be in esp format
  \usepackage{epsfig} %For pictures: screened artwork should be set up with an 85 or 100 line screen
\usepackage{graphicx}  \usepackage{epstopdf}%This is to transfer .eps figure to .pdf figure; please compile your paper using PDFLeTex or PDFTeXify.
 \usepackage[colorlinks=true]{hyperref}
\hypersetup{urlcolor=blue, citecolor=red}

\def\currentvolume{X}
 \def\currentissue{X}
  \def\currentyear{200X}
   \def\currentmonth{XX}
    \def\ppages{X--XX}
     \def\DOI{10.3934/fods.2020015}

\newcommand\be{\begin{equation}}
\newcommand\ee{\end{equation}}

\renewcommand\({\left(}
\renewcommand\){\right)}

\newcommand\Nx{{N_\mathrm{x}}}
\newcommand\Nu{{N_\mathrm{u}}}
\newcommand\Nz{{N_\mathrm{z}}}
\newcommand\Ny{{N_\mathrm{y}}}
\newcommand\Np{{N_\mathrm{p}}}
\newcommand\Ne{{N_\mathrm{e}}}

\newcommand{\Nq}{N_\mathrm{q}}
\newcommand\No{{N_0}}
\newcommand\Nexp{{N_\mathrm{exp}}}

\newcommand\bF{\mathbf{F}}
\newcommand\bS{\mathbf{S}}
\newcommand\bG{\mathbf{G}}
\newcommand\bQ{\mathbf{Q}}
\newcommand\bI{\mathbf{I}}
\newcommand\bH{\mathbf{H}}
\newcommand\bR{\mathbf{R}}
\newcommand\bB{\mathbf{B}}
\newcommand\Bxx{\mathbf{B}_\mathrm{xx}}
\newcommand\Bpx{\mathbf{B}_\mathrm{px}}
\newcommand\Bpp{\mathbf{B}_\mathrm{pp}}
\newcommand\Bxp{\mathbf{B}_\mathrm{xp}}
\newcommand\Bxy{\mathbf{B}_\mathrm{xy}}
\newcommand\Byy{\mathbf{B}_\mathrm{yy}}
\newcommand\bC{\mathbf{C}}
\newcommand\Cxx{\mathbf{C}_\mathrm{xx}}
\newcommand\Cpx{\mathbf{C}_\mathrm{px}}
\newcommand\Cpp{\mathbf{C}_\mathrm{pp}}
\newcommand\Cxp{\mathbf{C}_\mathrm{xp}}
\newcommand\Cxy{\mathbf{C}_\mathrm{xy}}
\newcommand\Cyy{\mathbf{C}_\mathrm{yy}}
\newcommand\bX{\mathbf{X}}
\newcommand\Xx{\mathbf{X}_\mathrm{x}}
\newcommand\Xp{\mathbf{X}_\mathrm{p}}
\newcommand\Xq{\mathbf{X}^\mathrm{q}}
\newcommand\bE{\mathbf{E}}
\newcommand\Ex{\mathbf{E}_\mathrm{x}}
\newcommand\Ep{\mathbf{E}_\mathrm{p}}
\newcommand\bY{\mathbf{Y}}

\newcommand\bw{\mathbf{w}}
\newcommand\wz{\mathbf{w}_\mathrm{z}}
\newcommand\wq{\mathbf{w}_\mathrm{q}}
\newcommand\bx{\mathbf{x}}
\newcommand\bu{\mathbf{u}}
\newcommand\bz{\mathbf{z}}
\newcommand\bp{\mathbf{p}}
\newcommand\by{\mathbf{y}}

\newcommand\brho{\boldsymbol \rho}
\newcommand\bdelta{\boldsymbol \delta}

\newcommand\bTheta{\boldsymbol \Theta}
\newcommand\bOmega{\boldsymbol \Omega}
\newcommand\bDelta{\boldsymbol \Delta}
\newcommand\bPi{\boldsymbol \Pi}
\newcommand\bXi{\boldsymbol \Xi}

\newcommand\T{^\top}

\newcommand\onehalf{\frac{1}{2}}
\newcommand\bzero{\mathbf{0}}
\newcommand\bone{\mathbf{1}}

\title[Online learning of both state and dynamics] %Use the shortened version of the full title
      {Online learning of both state and dynamics using ensemble Kalman filters}

\author[Marc~Bocquet, Alban~Farchi and Quentin~Malartic]{}


\email{marc.bocquet@enpc.fr}
\email{alban.farchi@enpc.fr}
\email{quentin.malartic@enpc.fr}

\thanks{$^*$ Corresponding author: Marc Bocquet}

\begin{document}
\maketitle

% Enter the first author's name and address:
\centerline{\scshape Marc Bocquet$^*$, Alban Farchi and Quentin Malartic}
\medskip
{\footnotesize
  % please put the address of the first author
  \centerline{CEREA, joint laboratory \'Ecole des Ponts ParisTech and EDF R\&D,}
 \centerline{Universit\'e Paris-Est, Champs-sur-Marne, France}
} % Do not forget to end the {\footnotesize by the sign }

\bigskip

% The name of the associate editor will be entered by an editorial staff
% "Communicated by the associate editor name" is not needed for special issue.
% \centerline{(Communicated by the associate editor name)}

%The abstract of your paper
 \begin{abstract}
     The reconstruction of the dynamics of an observed physical system as a surrogate model has been brought to the fore by recent advances in machine learning. To deal with partial and noisy observations in that endeavor, machine learning representations of the surrogate model can be used within a Bayesian data assimilation framework. However, these approaches require to consider long time series of observational data, meant to be assimilated all together.
  This paper investigates the possibility to learn both the dynamics and the state online, i.e. to update their estimates at any time, in particular when new observations are acquired. The estimation is based on the ensemble Kalman filter (EnKF) family of algorithms using a rather simple representation for the surrogate model and state augmentation. We consider the implication of learning dynamics online through (i) a global EnKF, (i) a local EnKF and (iii) an iterative EnKF and we discuss in each case issues and algorithmic solutions.
We then demonstrate numerically the efficiency and assess the accuracy of these methods using one-dimensional, one-scale and two-scale chaotic Lorenz models.
%% This is the abstract of your paper and it should not exceed \textbf{200} words.
\end{abstract}

%% \tableofcontents

\section{Introduction}

\subsection{Learning the dynamics of a model: A time-dependent variational problem}
\label{sec:var}

There has been a surge of studies that aim at reconstructing the dynamics of a physical system from its sole observation. The key output of these studies is a surrogate model meant to emulate the dynamical system. This trend has been triggered by the significant advances in machine learning (ML), and in particular neural networks (NNs), over the past decade. Chaotic systems are of particular interest because of their prevalence in geophysical flows,
but also because of their intrinsic instability and poor predictability making this endeavor a difficult one.
Several ML techniques have been tested when focusing on the reconstruction of chaotic dynamics: the analogs \cite{lguensat2017}, the projection of the dynamics resolvent onto nonlinear regressors (e.g., \cite{paduart2010, brunton2016}), echo state networks (e.g., \cite{pathak2018a}), and residual NNs to represent either the resolvent (e.g., \cite{dueben2018, scher2019, brajard2020a}) or underlying ordinary differential equations (ODEs) (e.g., \cite{fablet2018, long2018, bocquet2019b, bocquet2020}).
The ML techniques developed in these references are evaluated on chaotic low-order models such as the Lorenz-96 model (L96, \cite{lorenz1998}) or the two-scale Lorenz model (L05III, \cite{lorenz2005}).
Moreover, the NN and reservoir models in \cite{dueben2018, scher2019,weyn2019,arcomano2020} were tested on genuine meteorological fields.

In all these studies, the learning step is a variational problem which consists in optimizing a loss function. The latter crucially depends on the entire sequence of observations representing a long system trajectory, even though the observations can be exploited by batches, for instance when using stochastic optimization.

\subsection{A Bayesian framework combining data assimilation and machine learning}
In practice, most of these studies consider that the physical system is fully and noiselessly observed.
Some very weak noise can nonetheless be added to the observations for regularization of the learning scheme \cite{bishop1995}.
However, a Bayesian formalism that extends ML approaches, including NN representations, to the case of partial and noisy observations, has recently been developed.
Brajard et al. \cite{brajard2020a} have first shown how to combine ML with data assimilation (DA) to be able to process noisy and partial observations.
In their scheme, DA is used as an advanced space-time interpolation tool, which alternates with a refined estimation of the surrogate model through the optimization of the NN coefficients. This has been framed into a unifying Bayesian formalism in \cite{bocquet2020}, which allows to develop approximations and alternative algorithms.
The typical tool of the DA step is the ensemble Kalman smoother (EnKS), while the ML step relies on ML libraries such as TensorFlow or Pytorch.
Although this approach is theoretically scalable to high-dimensional systems, it is obviously more complex than the isolated ML step.
As emphasized in Section~\ref{sec:var}, the approach is offline as it requires the full observation dataset for the ML step.

Given a set of observations $\by_{0:K} =\left\{\by_k\right\}_{0 \le k \le K} \in {\mathbb R}^{(K+1) \times \Ny}$, a typical ML loss function is
\begin{align}
  \label{eq:ml-loss-function}
  \mathcal{J}(\bp) = \onehalf\sum_{k=1}^{K} \left\| \by_k-\bF^{k-1}(\bp, \by_{k-1}) \right\|^2
  + \mathcal{L}(\bp),
\end{align}
where $\bx \mapsto \bF^{k-1}(\bp, \bx) $ is the resolvent of the surrogate model integrated from $t_{k-1}$ to $t_k$, $\bp$ is the set of coefficients used
in the mathematical representation of the surrogate model (for instance the weights and biases of a NN), and $\mathcal{L}(\bp)$ is the regularization term
of the surrogate model, which, from a Bayesian standpoint, corresponds to a prior on the surrogate model.

In \eqref{eq:ml-loss-function}, the surrogate model is supposed to be autonomous, such that $\bp$ does not depend on time.
Generalizations are possible though, such as a slow evolution of the dynamics, or when considering a dynamical autonomous core parameterized by $\bp$ together with time-dependent parameters representing time-dependent forcings. However, these generalizations will not be addressed in this paper, so that the surrogate model will be assumed autonomous throughout the paper.

Equation \eqref{eq:ml-loss-function} is a limiting case of the more general cost function \cite{hsieh1998,abarbanel2018,bocquet2019b,bocquet2020}:
\begin{align}
  \label{eq:da-cost-function}
  \mathcal{J}(\bp,\bx_{0:K}) =&  \onehalf\sum_{k=0}^{K} \left\| \by_k-\bH_k(\bx_k) \right\|^2_{\bR^{-1}_k}  \nonumber \\
  & + \onehalf\sum_{k=1}^{K}  \left\| \bx_k-\bF^{k-1}(\bp,\bx_{k-1}) \right\|^2_{\bQ^{-1}_k}
  + \mathcal{L}(\bp, \bx_0) ,
\end{align}
where $\bx_{0:K} = \left\{\bx_k\right\}_{0 \le k \le K} \in {\mathbb R}^{(K+1) \times \Nx}$ is the unknown state trajectory, $\left\{ \bH_k \right\}_{0 \le k \le K}$ are the observation operators from ${\mathbb R}^\Nx$ to ${\mathbb R}^\Ny$. If need be, the observation space dimension $\Ny$
can be made time-dependent. The norm notation $\| \bx \|_\bG$ stands for the Mahalanobis distance $\sqrt{\bx\T\bG\bx}$.
This cost function is derived from Bayes' rule assuming Gaussian statistics for the errors: the observation errors have no bias and covariance matrices $\left\{\bR_k\right\}_{0 \le k \le K}$ and the model errors have no bias and covariance matrices $\left\{\bQ_k\right\}_{1 \le k \le K}$.
It is also assumed that these errors are uncorrelated in time and that model and observation errors are uncorrelated.
$\mathcal{L}(\bp, \bx_0)$ is the regularization term for the surrogate model coefficients $\bp$ and for the initial state $\bx_0$ of the trajectory, and can be derived from the prior distribution on $\bp$ and $\bx_0$.
The observation and model error statistics, i.e. $\bR_k$ and $\bQ_k$, are not considered part of the control variables, i.e. of the variable to be estimated.
This latter topic has been investigated in \cite{bocquet2020} and will not be dealt with in the present paper, as it would add another significant layer of complexity.

\subsection{Investigating theoretical aspects of online schemes}
\label{sec:online}

Our goal in this study is to investigate the possibility to learn both the state and the dynamics online, i.e. assimilate the new batches of observations, which are possibly sparse and noisy, when they arrive, and subsequently update the surrogate model and state estimates.

If the full observation dataset was available, the problem could be solved using the cost function \eqref{eq:da-cost-function}.
In practice with realistically noisy systems, one would have to use one of the algorithms described in \cite{brajard2020a, bocquet2020}. However, since we assume that the batches of observations
are gradually acquired, this requires the development of a sequential algorithm instead.
Our focus is on the theoretical and algorithmic aspects of the problem. Rasp \cite{rasp2020} has investigated a similar problem but focused on possible solutions for numerical weather prediction centers, in particular relying on the use of an existing imperfect model for the physical system. Our approach has more to do with the extension of known DA methods to ML problems.

A simple algorithm one can think of is the augmented state ensemble Kalman filter (EnKF), where the state variable $\bx$ is extended to incorporate the coefficients of the model representation $(\bx,\bp)$ following the principle introduced in \cite{jazwinski1970}.
The augmentation principle is elegant, simple and has already been used for parameterized model error or forcings in DA and inverse problems (see \cite{ruiz2013} for a review and references in the geosciences).

However, applying the augmented EnKF to the full model is numerically challenging.
It also raises many issues when considering more advanced schemes such as local EnKFs \cite{houtekamer2001, hamill2001, evensen2009} needed for high-dimensional estimation problems
and iterative EnKFs \cite{sakov2012, bocquet2012, sakov2018} meant to better deal with model nonlinearities.

The goal of this paper is to investigate algorithmically (i) the simple augmented EnKF for model and state reconstruction, as well as its advanced variants based on (ii) local EnKFs and on (iii) iterative EnKFs. Moreover, we want to numerically assess these algorithms on low-order chaotic models.

\subsection{Outline}

In Section~\ref{sec:representation}, we discuss choices to be made for the surrogate model mathematical representation, which may have an impact on the design of the augmented EnKF. In Section~\ref{sec:theory}, we define the augmented-state EnKF based on a ML surrogate model, introduce and develop the associated local EnKF, as well as the corresponding iterative EnKF. These methods are evaluated on the L96 and L05III one-dimensional low-order chaotic models in Section~\ref{sec:numerics}. Conclusions are drawn in Section~\ref{sec:conclusion}.

\section{Surrogate model representation}
\label{sec:representation}

Because the augmented EnKF is flexible, any adequate representation of the dynamics parameterized by a vector of coefficients $\bp$ can a priori be chosen as the surrogate model.
The main constraint is that the physical dynamics should project significantly onto the set of surrogate models generated by these parameters.
Another important constraint which is inherent to the choice of the EnKF and more fundamentally to the sequential approach, is that the number of coefficients $\Np$
in $\bp$ must be limited. This may rule out deep NNs with a large number of weights and biases, but more compact residual NNs based on convolutional layers could certainly be chosen as the surrogate model.
If such NN is implemented with TensorFlow or Pytorch, a bonus would be that the adjoint of the surrogate model could easily be obtained, since these tools come with automatic differentiation.
This is superfluous for the global and local EnKF since the adjoint is not required. By contrast, even though the adjoint is not strictly required, it could nevertheless be useful in the iterative EnKF or its local variant \cite{bocquet2016}, because the iterative EnKF piggybacks on a nonlinear variational analysis.

In the numerical applications of this study, we will choose the monomial representation described in Sections~2.1 and 2.2 of \cite{bocquet2019b}.
It is an ODE representation of the surrogate dynamics even though the physical model could either be governed by ODEs or partial differential equations (PDEs). The monomials are used to parameterize the flow rate, i.e. the equations of the surrogate model, which is later integrated in time using an appropriate numerical integration scheme (typically a fourth-order Runge--Kutta scheme) to build the resolvent $\bx_{k} \mapsto \bF^{k}(\bp,\bx_{k})$ between two time steps.
Note that this representation is equivalent to a NN, and can be either be implemented straightforwardly (as in \cite{bocquet2019b}) or using TensorFlow or Pytorch (as in \cite{bocquet2020}). Explicit details can be found in
\cite{bocquet2019b, bocquet2020}.

We make the assumption of locality of the dynamics, which reduces considerably the number of coefficients of the representation which then scales as the size of the system, see Section~2.2.1 of \cite{bocquet2019b}.
The mathematical description of the dynamics is local and only makes use of variables within a given stencil.
We further assume homogeneity of the system, see Section~2.2.2 of \cite{bocquet2019b}, i.e. translational invariance of the dynamics.
This assumption is less realistic but often satisfied by the dynamical part of the model, while forcing terms could realistically be spatially inhomogeneous.
In the end, the number of coefficients $\Np$ is considerably reduced by making these assumptions.
Both assumptions will be enforced hereafter, but homogeneity will be questioned to some extent in the following.

\section{Ensemble Kalman-based methods}
\label{sec:theory}

As explained in the introduction, Section~\ref{sec:online}, the augmented EnKF consists in extending the state vector $\bx \in \mathbb{R}^\Nx$ to
\begin{equation}
  \bz =  \left[ \begin{array}{c} \bx \\ \bp \end{array} \right] \in \mathbb{R}^\Nz,
\end{equation}
with $\Nz=\Nx+\Np$. The ensemble of the filter is made of such augmented vectors. Although the coefficients of $\bp \in \mathbb{R}^\Np$ are not observed, correlations between the state vector and the model parameters will implicitly be formed in the ensemble through the analysis. The best state/model couple will thrive in the forecast step and be subsequently favored in the analysis step by getting a larger contribution.

If the principle is well-known \cite{jazwinski1970, ruiz2013}, seeking a surrogate for the whole model is a bolder endeavor.
It also has important implications on the implementation of the local and of the iterative EnKFs, two fundamentally important variants of the EnKF.
In the rest of the paper, we will call EnKF-ML a variant of the EnKF with its state vector augmented with all surrogate model parameters.

\subsection{The ensemble Kalman filter}
\label{sec:enkf}

At time $t_k$, the basic EnKF-ML has its analysis step built on the cost function
\begin{equation}
  \label{eq:enkf-ml-cf}
  \mathcal{J}_k(\bz_k) = \onehalf\left\| \by_k-\bTheta_k(\bz_k) \right\|^2_{\bR^{-1}_k}  + \onehalf\left\| \bz_k-\bz_k^\mathrm{f} \right\|^2_{\bB^{-1}_k},
\end{equation}
where $\bz_k = [ \bx_k\T \quad \bp_k\T ]\T \in \mathbb{R}^\Nz$ and $\bTheta_k = [ \bH_k \quad \bzero ]$ is the augmented observation operator.
$\bB_k \in \mathbb{R}^{\Nz\times \Nz}$ is the error covariance matrix of $\bz_k$ estimated with the sample statistics of the augmented state ensemble\footnote{For the prior term in \eqref{eq:enkf-ml-cf} to make sense, one can assume $\bB_k$ to be invertible either by regularization, typically localization, or because its inverse can always be defined in the ensemble subspace.}.
The EnKF-ML has the same implementation as the EnKF albeit with the augmented state/model $\bz_k$.
In the numerical evaluation of Section~\ref{sec:numerics}, we choose for the EnKF-ML an implementation based on either the ensemble transform Kalman filter (ETKF, \cite{bishop2001,hunt2007}), or the ensemble square root Kalman filter (EnSRF, \cite{whitaker2002}).

For the state $\bx_k$, the forecast step from $t_k$ to $t_{k+1}$ is based on the application of the resolvent $\bx \mapsto \bF^k(\bp, \bx)$.
The forecast of the parameter is chosen to be persistence since we assumed the dynamics to be autonomous ($\bp_k$ is in principle constant in time). Because the reconstructed dynamics is certainly flawed,  with the academic exception of the perfect reconstruction of identifiable true dynamics, one has to account for model error in the forecast of the state,
for instance using the simple deterministic SQRT-CORE algorithm of \cite{raanes2015}, and/or using multiplicative inflation.

Because the augmented dynamics acting on $\bz_k$ are based on persistence for the parameters, it is clear that their Lyapunov spectrum will be that of $\bx \mapsto \bF^k(\bp_k, \bx)$ with the addition of $\Np$ neutral modes, i.e. of exponent $0$, due to the neutral dynamics of the model parameters\footnote{This would also be true for additive stochastic perturbations applied to the parameters.}. If the estimation of the physical model is accurate enough, the spectrum of  $\bx \mapsto \bF^k(\bp_k, \bx)$ should be close to that of the true dynamics.
Defining $\No$ as the dimension of the unstable and neutral subspace of the true dynamics, the dimension of the unstable and neutral subspace of the augmented dynamics should be about $\No+\Np$. As a consequence, following \cite{bocquet2017b,grudzien2018}, the size of the (centered) ensemble should at least satisfy
\begin{equation}
  \Ne \gtrapprox \No + \Np + 1,
\end{equation}
in order to avoid divergence without resorting to localization. From the theoretical and numerical results of \cite{bocquet2017a} on the impact of the neutral modes on the collapse of the forecast error covariance matrix on the unstable and neutral subspace, we expect that the filter would diverge for $\Ne \le \No $ and we expect that the accuracy of the filter would gradually improve in the range $\No +1 \le \Ne \le \No + \Np + 1$.
The filter's accuracy can however still improve when $\Ne > \No + \Np + 1$, as the need for inflation is increasingly reduced.

\subsection{The local ensemble Kalman filter}
\label{sec:lenkf}

In this section, we discuss of the implementation of a local EnKF-ML (LEnKF-ML). As explained above, whenever $\Ne < \No + \Np + 1$ localization was surmised to be beneficial, if not necessary.

\subsubsection{Localization and statistical homogeneity of the parameters}

The strategy chosen to make the filter local primarily depends on the nature of the surrogate model parameters $\bp$.
If the observed dynamical system is heterogeneous, the surrogate model representation should be heterogeneous as well, and $\bp$ should depend on space location. Hence those parameters are local.
In this case, the application of either local analysis/domain localization (DL) or covariance localization (CL) is in principle straightforward.
For DL, the update is performed locally for both the state and parameters, i.e. on the joint vector $\bz$. The updated ensemble can directly be formed
from these local updates. In this case, only local domains and possibly a tapering function for the observation precision matrix have to be specified as in standard EnKF DL. For CL, the localization matrix, to be used in the Schur product, can be defined using the physical distance between two state variables, two parameters, or a state variable and a parameter. One could also think of a more complex localization matrix with two localization lengths, one for the state variables and one for the parameters. Yet, one should check that the resulting correlation matrix is positive semi-definite.

These local parameter approaches are very appealing for the tentative LEnKF-ML. However, they would lead to severe underdetermination, especially in the sequential context of this study. There may be too many control variables to learn from for the observation batches. That is why we prefer to postpone the exploration of this approach and focus instead in the present paper in the case where the dynamical system is homogeneous. As discussed in the introduction, this implies that the surrogate model parameters could and should be chosen as global. If this may avoid the underdetermination hardship, this opens up to the intricate problem of the joint localization of the state variables and global parameters.

This important issue has been studied previously in \cite{aksoy2006, fertig2009, hu2010}. Their authors have proposed to make the global parameters local in the DL update step, followed by a spatial averaging of the local updated parameters to form the global parameters and be able to propagate the ensemble using these updated global parameters. By contrast, CL was chosen in \cite{koyama2010, ruckstuhl2018}, and localization was not applied in the global parameter space.
Indeed, it is difficult to choose a priori any correlation structure for the global parameters among themselves. Furthermore, it can be argued for the state/parameter covariances, that a global parameter should statistically be equally correlated to all state variables. Nonetheless, a tapering coefficient could be used for the state/parameter covariances. Ruckstuhl and Janji{\'c} \cite{ruckstuhl2018} empirically chose $\Nx^{-1}$ as tapering coefficient, and argue that this could make the localization matrix positive semi-definite (and hence a genuine correlation matrix).
Hereafter, we choose to work with the latter approach where there is no localization in parameter space but a tapering can be applied to the state/parameter covariances.

\subsubsection{Covariance localization}

In this section, we focus on the analysis at time $t_k$ and therefore we drop the time index $k$ for clarity.
The ensemble mean $\bar{\bz}$ and the normalized perturbation matrix $\bX$ of the ensemble $\left\{ \bz_i \right\}_{i=1, \ldots, \Ne}$ are defined by
\begin{equation}
  \bar{\bz} = \frac{1}{\Ne} \sum_{i=1}^\Ne \bz_i, \qquad \bX =  \left[ \frac{\bz_1 - \bar{\bz}}{\sqrt{\Ne-1}} \quad \frac{\bz_2 - \bar{\bz}}{\sqrt{\Ne-1}} \quad \ldots \quad \frac{\bz_\Ne - \bar{\bz}}{\sqrt{\Ne-1}} \right] .
\end{equation}
Let us split the background error covariances of $\bz$ according to the state and parameter subspaces:
\begin{equation}
  \bB = \left[ \begin{array}{cc} \Bxx & \Bxp \\ \Bpx & \Bpp \end{array} \right] .
\end{equation}
We assume that $\bB$ has already been localized. One would typically write $\bB = \bC \circ \left(\bX\bX\T\right)$ where  $\circ$ is the Schur product and $\bC$ is an admissible localization matrix, which can be written block-wise
\begin{equation}
    \bC = \left[ \begin{array}{cc} \Cxx & \Cxp \\ \Cpx & \Cpp \end{array} \right].
\end{equation}

In particular, this implies that $\Bxx$ must be invertible.
Given that the parameters are not observed, the state and parameter update reads:
\begin{subequations}
  \label{eq:state-update}
\begin{align}
  \bar{\bx}^\mathrm{a} &= \bar{\bx}^\mathrm{f} + \Bxx\bH\T \left( \bR+\bH\Bxx\bH\T \right)^{-1}\left( \by-\bH \bar{\bx}^\mathrm{f} \right), \\
    \bar{\bp}^\mathrm{a} &=\bar{\bp}^\mathrm{f} + \Bpx\bH\T \left( \bR+\bH\Bxx\bH\T \right)^{-1}\left( \by-\bH \bar{\bx}^\mathrm{f} \right).
\end{align}
\end{subequations}
Let us note that the parameter update can be written as
\begin{equation}
  \bar{\bp}^\mathrm{a} = \bar{\bp}^\mathrm{f} + \Bpx\Bxx^{-1} \left( \bar{\bx}^\mathrm{a}-\bar{\bx}^\mathrm{f} \right),
\end{equation}
which confirms that, given that the parameters are not observed, their update can be computed by a simple linear regression from the state update, using the prior statistics.
In practice, one first solves the system $\Bxx \bdelta = \bar{\bx}^\mathrm{a}-\bar{\bx}^\mathrm{f}$ for $\bdelta \in {\mathbb R}^\Nx$, and then computes $\bar{\bp}^\mathrm{a} = \bar{\bp}^\mathrm{f} + \Bpx\bdelta $. The system can be solved with a linear solver without having to explicitly compute the Schur product of $\Bxx$ (see, e.g., \cite{farchi2019}).

The perturbation update based on CL has the theoretical form \cite{sakov2011}:
\begin{equation}
  \label{eq:leftT}
  \bX^\mathrm{a} = \left( \bI + \bB\bTheta\T\bR^{-1}\bTheta \right)^{-\onehalf} \bX^\mathrm{f},
\end{equation}
where $\bI$ is the identity matrix, and $\bTheta$ stands for the tangent linear of the augmented observation operator used in \eqref{eq:enkf-ml-cf}.
A careful definition of the square root matrix used in this formula can be found, e.g., in Section~2.3.1 of \cite{bocquet2019a}.
This update can in practice be computed resorting to augmented ensemble techniques \cite{bishop2017,farchi2019}, which is of special interest for the assimilation of nonlocal observations.
Equation~\eqref{eq:leftT} can be expanded into state and parameter blocks. We define $\Xx \in \mathbb{R}^{\Nx\times \Ne}$ and $\Xp \in \mathbb{R}^{\Np\times \Ne}$ as the state and parameter block matrices of $\bX$, respectively, i.e. the state and parameter normalized perturbations matrices. Then expanding \eqref{eq:leftT} and using (25) in \cite{bocquet2016}, we obtain
\begin{equation}
  \label{eq:leftT-exp}
  \bX^\mathrm{a} = \left[ \begin{array}{c}
      \left( \bI + \Bxx\bOmega \right)^{-\onehalf} \Xx^\mathrm{f} \\
      \Xp^\mathrm{f} - \Bpx\bOmega\left( \bI + \Bxx\bOmega
      + \left(\bI + \Bxx\bOmega \right)^{\onehalf}\right)^{-1} \Xx^\mathrm{f}
    \end{array}
    \right],
\end{equation}
with the notation $\bOmega = \bH\T\bR^{-1}\bH$.
The first block, i.e. $\Xx^\mathrm{a}$ of \eqref{eq:leftT-exp} can be computed using an augmented ensemble technique.
Using (25) in \cite{bocquet2016}, it can be shown that $\Xp^\mathrm{a}$ can also be written:
\begin{align}
  \label{eq:par-pert-update}
  \Xp^\mathrm{a} &=  \Xp^\mathrm{f} - \Bpx\Bxx^{-1}\Bxx\bOmega\left( \bI + \Bxx\bOmega
  + \left(\bI + \Bxx\bOmega \right)^{\onehalf}\right)^{-1} \Xx^\mathrm{f} \nonumber \\
  & = \Xp^\mathrm{f} + \Bpx\Bxx^{-1} \left\{ \left( \bI + \Bxx\bOmega \right)^{-\onehalf} - \bI \right\} \Xx^\mathrm{f} \nonumber \\
  & = \Xp^\mathrm{f} + \Bpx\Bxx^{-1}\left( \Xx^\mathrm{a} - \Xx^\mathrm{f} \right).
\end{align}
Hence the updated perturbations of the parameters can be obtained by a linear regression using the prior statistics, with the updated state perturbations.
Hence, computing \eqref{eq:par-pert-update} alternately consists of (i) solving the linear system $\Bxx \bDelta = \Xx^\mathrm{a} - \Xx^\mathrm{f}$ for $\bDelta \in {\mathbb R}^{\Nx \times \Ne}$ and (ii) computing $\Xp^\mathrm{a} = \Xp^\mathrm{f} + \Bpx \bDelta$.

The update in parameter space can conveniently be summarized using the ensemble blocks.
The ensemble matrix $\bE = [ \bz_1 \quad \bz_2 \quad \cdots \quad \bz_\Ne ]$ is decomposed into $[ \Ex\T \quad \Ep\T ]\T$ for the state and parameter ensembles. The mean state and perturbations update for the parameters can then be compactly written:
\begin{equation}
  \label{eq:ensemble-update-cl}
  \Ep^\mathrm{a} = \Ep^\mathrm{f} + \Bpx\Bxx^{-1} \left( \Ex^\mathrm{a}-\Ex^\mathrm{f} \right),
\end{equation}
which can be computed (i) solving the linear system $\Bxx \bDelta = \Ex^\mathrm{a} - \Ex^\mathrm{f}$ and (ii) updating $\Ep^\mathrm{a} = \Ep^\mathrm{f} + \Bpx \bDelta$.

As a conclusion, for an LEnKF-ML based on CL, one would first (i) update the state space part of the ensemble and then (ii) update
the parameter space part of the ensemble using the state ensemble incremental update $\Ex^\mathrm{a}-\Ex^\mathrm{f}$ and \eqref{eq:ensemble-update-cl} with
$\Bxx= \Cxx \circ \left(\Xx^\mathrm{f}\left(\Xx^\mathrm{f}\right)\T\right)$ and $\Bpx = \zeta \Xp^\mathrm{f}\left(\Xx^\mathrm{f}\right)\T$, where $\zeta$ is the tapering state/parameter coefficient.

\subsubsection{Domain localization}

There are several variants of DL \cite{houtekamer2001,ott2004}. In this section, we use the algorithm of the local ETKF (LETKF) as formalized in \cite{hunt2007}, with most of its matrix algebra performed in ensemble space.
We describe what could be the implementation of a local EnKF-ML based on the LETKF algorithm and global parameters.

Let us define the observation anomaly: $\bY = \bTheta\bX^\mathrm{f} = \bH\Xx^\mathrm{f}$, where $\bTheta$, $\bH$ are the tangent linear of the observation operators introduced in \eqref{eq:enkf-ml-cf}. They can alternatively be computed with the secant method using the ensemble and the nonlinear operators.
The state and parameter update \eqref{eq:state-update}, but where $\bB$ corresponds now to the raw sample covariance matrix (i.e. without tapering), becomes
\begin{subequations}
  \label{eq:state-update-ens}
  \begin{align}
    \label{eq:state-update-ens-x}
  \bar{\bx}^\mathrm{a} &= \bar{\bx}^\mathrm{f} + \Xx^\mathrm{f} \bw^\mathrm{a}, \qquad \bw^\mathrm{a} =\left( \bI+\bY\T \bR^{-1}\bY \right)^{-1}\bY\T\bR^{-1}\left( \by-\bH \bar{\bx}^\mathrm{f} \right),
  \\
  \label{eq:state-update-ens-p}
    \bar{\bp}^\mathrm{a} &=\bar{\bp}^\mathrm{f} + \Xp^\mathrm{f} \bw^\mathrm{a} .
\end{align}
\end{subequations}
While it is clear that \eqref{eq:state-update-ens-x} can be implemented as a traditional LETKF state update, a local domain implementation
of \eqref{eq:state-update-ens-p} is pointless. Indeed, $\Xp^\mathrm{f}$ has no connection to state space so that one would not know
how to apply a local weight update of $\bw^\mathrm{a}$ to it.

Let us assume that the null space of $\bX^\mathrm{f}$ is spanned by $\bone=[1 \quad  1 \quad \ldots \quad 1]\T \in \mathbb{R}^\Ne$,
so that, by the rank theorem $\Ne \le \Nx + 1$, which is a realistic assumption for the LEnKF-ML.
As a consequence, $\Xx^\mathrm{f}\bone = \bzero$ and $\Xp^\mathrm{f}\bone = \bzero$.
Hence, we have
\be
\label{eq:rank}
\Xp^\mathrm{f} \left(\Xx^\mathrm{f}\right)^\dagger \Xx^\mathrm{f} = \Xp^\mathrm{f} \(\bI - \frac{\bone \bone\T}{\Ne} \) = \Xp^\mathrm{f},
\ee
where $\left(\Xx^\mathrm{f}\right)^\dagger$ is the Moore-Penrose inverse of $\Xx^\mathrm{f}$. Then, \eqref{eq:state-update-ens-p} can be transformed into
\begin{equation}
  \label{eq:state-update-ens-p2}
  \bar{\bp}^\mathrm{a} = \bar{\bp}^\mathrm{f} + \Xp^\mathrm{f} \left(\Xx^\mathrm{f}\right)^\dagger \Xx^\mathrm{f} \bw^\mathrm{a}
  = \bar{\bp}^\mathrm{f} + \Xp^\mathrm{f}\left(\Xx^\mathrm{f}\right)^\dagger \left(\bar{\bx}^\mathrm{a}- \bar{\bx}^\mathrm{f}\right).
\end{equation}
With DL enforced, $\Xx^\mathrm{f} \bw^\mathrm{a}$ in \eqref{eq:state-update-ens-x} is computed following the traditional LETKF and $\bar{\bx}^\mathrm{a}$ appearing in \eqref{eq:state-update-ens-p2} corresponds to the LETKF state update.

As for the perturbation update, the right-transform update is:
\begin{equation}
  \label{eq:rightT}
  \bX^\mathrm{a} =  \bX^\mathrm{f} \left(\bI + \bY\T\bR^{-1}\bY\right)^{-\onehalf},
\end{equation}
for the augmented state and can be expanded as
\begin{subequations}
    \label{eq:rightT-exp}
\begin{align}
  \label{eq:rightT-exp-x}
  \Xx^\mathrm{a} &= \Xx^\mathrm{f}\left(\bI + \bY\T\bR^{-1}\bY\right)^{-\onehalf}, \\
  \label{eq:rightT-exp-p}
 \Xp^\mathrm{a} &=      \Xp^\mathrm{f}\left(\bI + \bY\T\bR^{-1}\bY\right)^{-\onehalf}.
\end{align}
\end{subequations}
The DL LEnKF-ML state perturbation update as given by \eqref{eq:rightT-exp-x}
hence follows the traditional LETKF state perturbation update.
As for the parameter perturbation update, \eqref{eq:rightT-exp-p}
can be written (still assuming \eqref{eq:rank}):
\begin{equation}
  \label{eq:parameter-perturbation-update}
  \Xp^\mathrm{a} = \Xp^\mathrm{f}\left(\Xx^\mathrm{f}\right)^\dagger \Xx^\mathrm{f} \left(\bI + \bY\T\bR^{-1}\bY\right)^{-\onehalf} =
  \Xp^\mathrm{f}\left(\Xx^\mathrm{f}\right)^\dagger \Xx^\mathrm{a},
\end{equation}
where $\Xx^\mathrm{a}$ has been obtained using the LETKF perturbation update.

Similarly to CL, it is then straightforward to show from Eqs.~(\ref{eq:state-update-ens-p2}, \ref{eq:parameter-perturbation-update}) that those results can be summarized into a compact ensemble update formula for the parameters:
\begin{equation}
  \label{eq:ensemble-update-dl}
  \Ep^\mathrm{a} = \Ep^\mathrm{f} + \Xp^\mathrm{f}\left(\Xx^\mathrm{f}\right)^\dagger \left( \Ex^\mathrm{a}-\Ex^\mathrm{f} \right),
\end{equation}
where $\Ex^\mathrm{a}$ has been obtained from the ensemble update
of the traditional LETKF. This could be computed by first solving the over-constrained linear system $\Xp^\mathrm{f} \boldsymbol{\Delta} = \Ex^\mathrm{a}-\Ex^\mathrm{f}$ and then computing $\Ep^\mathrm{a} = \Ep^\mathrm{f} + \Xp^\mathrm{f}\boldsymbol{\Delta}$.
Finally, note that it is possible to taper the state/parameter error covariances and
modify \eqref{eq:ensemble-update-dl} into
\begin{equation}
 \label{eq:ensemble-update-dl-2}
 \Ep^\mathrm{a} = \Ep^\mathrm{f} + \zeta\Xp^\mathrm{f}\left(\Xx^\mathrm{f}\right)^\dagger \left( \Ex^\mathrm{a}-\Ex^\mathrm{f} \right),
\end{equation}
where $\zeta$ is the tapering coefficient. This would make sense to do so in the DL context since such tapering has not been accounted for yet in the algorithm.

More generally, this two-step DL-based EnKF with global parameters can be seen as an efficient alternative to the approach of \cite{aksoy2006, fertig2009, hu2010}, without the approximate averaging step. As a downside, one must carefully choose the tapering coefficient $\zeta$.

\subsubsection{Tapering of the state/parameter covariances}
\label{sec:tapering}

In this section, arguments are given for the use of a tapering coefficient of the state/parameter covariances.
We follow the discussion of \cite{ruckstuhl2018} in their Section~3.1, and slightly generalize their derivation. The main argument of \cite{ruckstuhl2018} is that the localization matrix $\bC$ should be positive definite. We note, however, that this is a sufficient condition for the regularized $\bB$ to be positive definite, not a necessary condition. In contrast to \cite{ruckstuhl2018}, all parameters are considered here.
We have $\Cxx=\brho$, the localization matrix in state space, which is assumed to be positive definite.
As explained above, we choose not to apply localization in parameter space, such that $\Cpp = \bPi_\mathrm{pp}$, the matrix full of $1$ of size $\Np \times \Np$. The cross state/parameter correlations which accounts for the tapering reads $\Cpx= \Cxp\T = \zeta \bPi_\mathrm{px}$
where $\bPi_\mathrm{px}$ is the matrix full of $1$ of size $\Np \times \Nx$.
Since $\Cxx=\brho > \bzero$ is positive definite ($>$ stands for the Loewner order of symmetric matrices), the full localization matrix $\bC$ is positive definite if and only if the following Schur complement $\bS$ is positive definite (Theorem 7.7.7 in \cite{horn2013}):
\begin{equation}
  \bS = \Cpp-\Cpx \Cxx^{-1} \Cxp
  = \bPi_\mathrm{pp} - \zeta^2 \bPi_\mathrm{px}\brho^{-1}\bPi_\mathrm{xp}  > \bzero,
\end{equation}
where $\bPi_\mathrm{xp}=\bPi_\mathrm{px}\T$.
Note that Ruckstuhl and Janji{\'c} \cite{ruckstuhl2018} chose to consider the other Schur complement defined in state subspace.
For any $\bp \in \mathbb{R}^\Np$, we have
\begin{equation}
  \bp\T \bS \bp = \sigma^2\left(1 - \zeta^2 \bone\T\brho^{-1} \bone\right),
\end{equation}
where $\sigma=\sum_{i=1}^\Ne p_i$ and $\bone = [1 \quad  1 \quad \ldots \quad 1]\T \in {\mathbb R}^\Nx$.
If $\lambda_\mathrm{min}>0$ is the smallest eigenvalue of the positive definite $\brho$,
we have
\begin{equation}
  \bp\T \bS \bp \ge \sigma^2\left(1 - \zeta^2\Nx \lambda_\mathrm{min}\right).
  \end{equation}
Hence, a necessary condition for $\bS$, and hence $\bC$, to be positive semi-definite is
\begin{equation}
  \label{eq:bound}
  \zeta < \sqrt{\frac{\lambda_\mathrm{min}}{\Nx}}.
\end{equation}
For a homogeneous regular (non-random) positive $\brho$, $\lambda_\mathrm{min}$ is expected to scale like $\Nx^{-1}$.
%% \todo{To be justified.}
Hence, from \eqref{eq:bound} an upper bound for $\zeta$
should scale like $g(\Ne)/\Nx$ with increasing $\Nx$, where $g(\Ne)$ is some function of $\Ne$.
The choice $\zeta = \Nx^{-1}$ was astutely proposed on empirical grounds in \cite{ruckstuhl2018}.

\subsection{The iterative ensemble Kalman filter}
\label{sec:ienkf}

Developing a local EnKF-ML is certainly the most critical task for the scalability of the approach.
By contrast, developing an iterative EnKF-ML is of theoretical interest, in particular when the time interval between updates is large
enough so that the nonlinearity of the true model significantly emerges.
In such circumstances, the performance of the EnKF degrades and it has to be replaced by its iterative variant, the IEnKF \cite{sakov2012,bocquet2012}.
An IEnKF-ML is an intriguing subject, because the nonlinear dynamical system has to be estimated along with the state trajectory.
Joint state and parameter estimation with an IEnKF was explored in \cite{bocquet2013}, but the parameters were only constraining the linear part of the dynamics. The IEnKF was used to retrieve boundary conditions of a computational fluid dynamics problem in \cite{defforge2019}, but the dynamics were assumed stationary.

In an IEnKF-ML, not only is state augmentation used as in the EnKF-ML but, by contrast, it is also optimized on in the analysis step of the method which is of variational nature.
For the sake of simplicity, we will not use localization with the IEnKF-ML, although localized variants are possible (see \cite{bocquet2016,sakov2018}).
The cost function that underlies the analysis of the IEnKF-ML is defined over a lag-one DA window (DAW) $[t_{0},t_{1}]$.

Because the surrogate model is likely to be flawed, especially in the first cycles of the (I)EnKF-ML sequential runs, model error should be accounted for, either by multiplicative inflation or by addition of stochastic noise on the state. Hence, a natural framework for the IEnKF-ML is the IEnKF-Q \cite{sakov2018}, i.e. an IEnKF that rigorously accounts for additive model error.
The IEnKF-ML cost function, based on the generalization of both the cost function \eqref{eq:enkf-ml-cf} of the EnKF-ML
and the cost function of the IEnKF-Q defined in (2b) of \cite{sakov2018}, reads
\begin{equation}
  \label{eq:ienkf-ml-cf}
  \mathcal{J}(\bz_0, \bz_1) = \onehalf\left\| \by_1-\bTheta_1(\bz_1) \right\|^2_{\bR_1^{-1}}
+ \onehalf\left\| \bz_1-\mathcal{M}_{0}(\bz_0) \right\|^2_{\bXi_1^{-1}}
  + \onehalf\left\| \bz_0-\bar{\bz}_0^\mathrm{f} \right\|^2_{\bB_0^{-1}},
\end{equation}
where
\begin{equation}
  \mathcal{M}_{0}(\bz_0) = \left[ \begin{array}{c} \bF^{0}(\bz_0) \\ \bp_0 \end{array} \right]
\end{equation}
with $\bF^{0}(\bz_0)$ standing for $\bF^{0}(\bp_0, \bx_0)$, and
assuming the persistence model for the surrogate model parameters. Note that depending on the definition of the model noise statistics $\bXi_1$, stochastic perturbations are also possible for the propagation of the surrogate model parameters, even though the deterministic parameter evolution model is persistence.

Thanks to the augmented state trick, \eqref{eq:ienkf-ml-cf}  has exactly the same form as that of the IEnKF-Q, so that we closely follow the derivation of its analysis step.
We introduce the state model perturbation matrix as $\Xq_1 \in \mathbb{R}^{\Nz \times \Nq}$ and it is defined through $\bXi_1 = \Xq_1\left(\Xq_1\right)\T$, where $\Nq$ is the number of centered model error perturbations.
Sakov et al. \cite{sakov2018} discussed of the relevance of this factorization.
The cost function can be written in ensemble space through the parameterization:
\begin{subequations}
  \label{eq:ienkf-ml-constraints}
\begin{align}
  \bz_0 &= \bar{\bz}^\mathrm{f}_0 + \bX_0\wz, \\
  \label{eq:ienkf-ml-constraints-1}
   \bz_1 &= \mathcal{M}_0(\bz_0) + \Xq_1\wq .
\end{align}
\end{subequations}
Denoting $\bw = [\wz\T \quad \wq\T ]\T \in \mathbb{R}^{\Ne+\Nq}$ and following \cite{sakov2018}, we obtain a very compact cost function in ensemble space:
\begin{equation}
  \label{eq:ienkf-ml-cf-3}
  \mathcal{J}(\bw) = \onehalf\left\| \by_1-\bTheta_1(\bz_1) \right\|^2_{\bR_1^{-1}}
  + \onehalf\left\| \bw\right\|^2 ,
\end{equation}
under the constraints \eqref{eq:ienkf-ml-constraints}. Because it has the same mathematical structure as (8) of \cite{sakov2018},
the algorithm of the IEnKF-ML is formally the same as that of the IEnKF-Q, see Algorithm 1 of \cite{sakov2018}, with the state vectors being replaced with state/parameter augmented vectors.

Note that, for the sake of simplicity, we could choose $\bXi^{-1}_1$ to be of the form:
\begin{equation}
  \bXi_1^{-1} = \left( \begin{array}{cc} \bQ_1^{-1} & \bzero \\ \bzero & \bzero \end{array} \right) ,
\end{equation}
i.e. the parameter propagation is persistence only, without additional stochastic perturbations.
In that case, \eqref{eq:ienkf-ml-cf} reads
\begin{equation}
  \label{eq:ienkf-ml-cf-2}
  \mathcal{J}(\bz_0, \bz_1) = \onehalf\left\| \by_1-\bH_1(\bx_1) \right\|^2_{\bR_1^{-1}}
+ \onehalf\left\| \bx_1-\bF^{0}(\bz_0) \right\|^2_{\bQ_1^{-1}}
  + \onehalf\left\| \bz_0-\bar{\bz}_0^\mathrm{f} \right\|^2_{\bB_0^{-1}},
\end{equation}
the model error perturbation matrix becomes $\Xq_1 \in \mathbb{R}^{\Nx \times \Nq}$ such that $\bQ_1 = \Xq_1\left(\Xq_1\right)\T$, and \eqref{eq:ienkf-ml-constraints-1} should read
\begin{equation}
   \bx_1 = \bF^{0}(\bz_0) + \Xq_1\wq .
\end{equation}
Finally, \eqref{eq:ienkf-ml-cf-3} could be re-written:
\begin{equation}
  \label{eq:ienkf-ml-cf-4}
  \mathcal{J}(\bw) = \onehalf\left\| \by_1-\bH_1(\bx_1) \right\|^2_{\bR_1^{-1}}
  + \onehalf\left\| \bw\right\|^2 .
\end{equation}

We also remark that the decoupling of $\wz$ and $\wq$ when the observation operator is linear, which was put forward and explained in Section~3 of \cite{sakov2018} and Appendix B of \cite{fillion2020}, also happens for the IEnKF-ML.

Let us now discuss two appealing properties of the IEnKF-ML, which may help scale it up to high dimensional systems.

The IEnKF is a very powerful nonlinear filtering technique. But it has a numerical cost due to the propagation of the ensemble within the DA window for each iteration of the nonlinear variational analysis. The average number of iterations can be kept quite small in a cycled DA run, as opposed to a 4D-Var, but the number of ensemble propagations may easily be twice that of a traditional EnKF. By contrast, the propagation of the ensemble within the variational analysis of the IEnKF-ML is carried out using the surrogate model, which is likely to be numerically much cheaper especially when using efficient highly-parallelized, ML libraries.

As opposed to the EnKF-ML, the adjoint of the surrogate model is explicitly or implicitly used in the IEnKF-ML.
In the Algorithm 1 of \cite{sakov2018}, it is evaluated using finite-differences or the secant method, and a simple transposition in the ensemble subspace.
This is the approach used in ensemble variational DA techniques (see Section~4.5 of \cite{carrassi2018}), and which was also advocated in the ML context in the absence of an adjoint model \cite{kovachki2019}.
If the surrogate model is implemented in TensorFlow or Pytorch, an efficient adjoint can be obtained, and would possibly make the IEnKF-ML more accurate, especially if localized or if a static background is additionally used in the analysis. Moreover, this would avoid the need for ensemble propagations within the DA window.

\section{Numerical results}
\label{sec:numerics}

In this section, we test the global and local EnKF-ML as well as the IEnKF-ML,
mostly with the low-order chaotic model L96 but also with the L05III model.
Both models are very commonly employed to test new DA schemes so that they are well suited for this study.
The true L96 model lies in the set of all achievable surrogate models by the representation described in Section~\ref{sec:representation}. Hence, the true dynamics could theoretically be reconstructed.
By contrast, the true L05III model, which has two scales of motion, cannot be accurately represented by the surrogate model representation chosen in Section~\ref{sec:representation} since it only accounts for a single scale of motion (no nested submodel).

\subsection{Lorenz-96}

The L96 model is defined by a set of ODEs over a periodic domain with variables indexed by $n=0,\ldots,N_x-1$:
\begin{equation}
  \frac{{\mathrm{d}}x_n}{{\mathrm{d}}t} = (x_{n+1}-x_{n-2})x_{n-1}-x_n+F ,
\end{equation}
where $N_x=40$, $x_{N_x}=x_0$, $x_{-1} = x_{N_x-1}$, $x_{-2}=x_{N_x-2}$, and $F=8$.
This model is an idealized representation of a one-dimensional latitude band of the Earth atmosphere.
The dimension of the unstable and neutral subspace is $\No=14$.
The truth run of the L96 model is integrated using the fourth-order Runge--Kutta (RK4) scheme with a time step of $\delta t=0.05$.

The surrogate model to be used has a one-to-one correspondence with all L96 variables $\left\{ x_n \right\}_{n=0, \ldots, N_x-1}$,
so that it can exhaustively represent the dynamics of L96. The stencil of the surrogate model, as defined in Section~\ref{sec:representation}, is chosen to be $L=2$ (i.e. contains $2L+1=5$ local variables).

\subsection{Lorenz-05III}

The two-scale Lorenz model L05III is given by the following two-scale set of ODEs:
  \begin{subequations}
    \label{eq:l05III}
    \begin{align}
      \frac{\mathrm{d}x_n}{\mathrm{d}t} &= \psi^+_n(\bx)+F-h\frac{c}{b}\sum_{m=0}^{9} u_{m+10n}, \\
      \frac{\mathrm{d}u_m}{\mathrm{d}t} &= \frac{c}{b} \psi^-_m(b\bu) + h\frac{c}{b} x_{m/10}, \\
      \psi_n^\pm (\bx) &= x_{n\mp 1}(x_{\pm1}-x_{n\mp 2})-u_n,
    \end{align}
  \end{subequations}
  for $n=0,\ldots,N_x-1$ with $N_x=36$, and $m=0,\ldots,N_u-1$ with $N_u=360$. The indices are defined periodically over their domains and $m/10$ stands for the integer division of $m$ by $10$. The other parameters are set to their original values: $c=10$ for the time-scale ratio, $b=10$ for the space-scale ratio, $h=1$ for the coupling between the scales, and $F=10$ for the forcing.
When uncoupled ($h=0$), the dimension of the unstable and neutral subspace of the coarse modes model compartment is $\No=13$ (for a thorough analysis of this dynamical system, see \cite{carlu2019}).
The stencil of the surrogate model is chosen to be $L=2$.
The vector $\bu$ represents unresolved scales and hence model error when only considering the slow variables $\bx$. It is integrated with an RK4 scheme and the time step $\delta t=0.005$ since it is stiffer than the L96 model.

The surrogate model has a one-to-one correspondence with the coarse scale variables $\left\{ x_n \right\}_{n=0, \ldots, N_x-1}$ only, so that it can only represent an approximation of the dynamics of L05III.

\subsection{Objective and metric}

The objective is to estimate both the state trajectory and the dynamics as a set of ODEs.
For the global and local EnKF-ML, and for the IEnKF-ML, we make a selection of what we believe to be relevant and revealing experiments since
this generalized DA/ML problem has many degrees of freedom in its setup, even more than typical DA experiments.

The metric is the average (over time) root mean square error of the state analysis (RMSE).
We do not report a metric for the dynamics as this is not relevant for L05III and because, in such a sequential context, the fitness of the surrogate dynamics will manifest itself in the state variable RMSE.
The RMSE will be averaged over $\Nexp$ long data assimilation runs for the error statistics to be reliable.

For each of the DA experiments, the setup will be specified. They correspond to, or are variations of classical DA configurations for the L96 and L05III models.

\subsection{Numerical complexity}
The numerical cost of the EnKF-ML depends primarily on the ensemble size, which itself depends on many different factors and choices: state space dimension of the surrogate dynamics; complexity, stability and symmetries of the true dynamics and its observation operator; efficacy of the surrogate model representation, etc.

However, it is clear that its numerical cost is mainly driven by the number of model parameters to estimate.
%, either global for the EnKF-ML or local for the LEnKF-ML.
Indeed, as demonstrated in Section~\ref{sec:theory}, the ensemble size is an affine function of the number of parameters. It also strongly depends on the choice of the surrogate model representation and whether it leverages on efficient machine learning libraries such as TensorFlow or PyTorch.
For instance, using the setup of the following sections, an EnKF-ML DA run applied to the observations of the L96 model would be ten times slower than an EnKF DA run knowing the true dynamics. But an EnKF-ML DA run applied to the observations of coarse modes of the L05III model would only be twice slower than an EnKF DA run knowing the true dynamics since the surrogate state vector is of lower dimensionality.

\subsection{The EnKF-ML}
\label{sec:enkf-ml}

In this section, we study numerically the basic EnKF-ML, which is based on an ETKF, without either localization or nonlinear iterations.
Because there are so many hyperparameters to consider such as ensemble size, frequency and sparsity of observations, inflation on the state or the parameters, model error noise on the state, etc, we have performed several sensitivity studies to freeze a few of these hyperparameters, hoping not to lose much in terms of accuracy of the filters in the process. The most critical ones focus on the ensemble initialization and on the ensemble inflation schemes.

\subsubsection{Initialization of the ensemble}

The success of an EnKF-ML run critically depends on the initialization of the ensemble of model parameters, for both low-order models under scrutiny.
Such sensitivity to the initial ensemble is not surprising since both state and model are quite uncertain at the beginning of the DA run.
By contrast, with such simple models and when the model is perfectly known, i.e. with traditional DA, the performance of a long DA run barely depends on the initialization of the ensemble, provided its spread is consistent with that of the error in the state initial condition.

From \cite{bocquet2017a, bocquet2017b}, we expect that with an ensemble size greater than $\No+\Np$, the RMSE over a long run should not depend on the initialization.
We have experimented with the standard deviation of the initial model parameters estimate and found a strong dependence of the average EnKF-ML RMSE when $\Ne \le \No+\Np$.
This is consistent with our theory. However, for $\Ne \ge \No+\Np+1$, we found a residual but significant dependence of the RMSE on the ensemble initialization, which contradicts the theoretical expectations. This is likely due to the use of a uniform multiplicative inflation to counteract model error, i.e. the same scalar inflation is applied at every time step. A strong inflation factor is required in the burn-in of the DA run when the uncertainty on the model parameters estimate is significant. But since this inflation is not adaptive, it has a negative impact on the average RMSE of the experiment when, on the long term, the model is better estimated.
Hence an adaptive inflation scheme is needed. This again contrasts with the traditional EnKF applied to simple homogeneous low-order models, such as those considered here, where a uniform inflation often yields optimal RMSEs.

Hence, in the following, but for the global EnKF-ML only where such sensitivity is critical, we shall append the adaptive inflation scheme developed in \cite{raanes2019} on top of the EnKF-ML. This scheme finely accounts for both model error and sampling error (that are generated by the model nonlinearities) by combining an efficient model error estimation scheme, essentially the one described in \cite{miyoshi2011}, with the finite-size scheme \cite{bocquet2011}. If adopting such an hybrid adaptive scheme is theoretically needed, this also has the key practical advantage that no experimentation on the inflation is required, which strongly decreases the number of DA experiments of our investigations.

One drawback is that only one adaptive scalar multiplicative inflation is estimated for the augmented state, whereas the augmented space is by construction inhomogeneous.
We have experimented with a dual inflation scheme where a uniform inflation is applied to the state space while another is applied to the model parameter space.
However, we found a very limited advantage in terms of RMSE of using a dual inflation over a single one in the augmented space, so that we do not resort to a dual inflation in the rest of this paper.

Let us illustrate the dependence on the initial ensemble using the EnKF with the hybrid adaptive inflation scheme, and applied to the L96 model.

We perform $\Nexp=100$ identical experiments with as many different random seeds, with an ensemble size in the range $\Ne=[12, 52]$. The ensemble of runs allows to compute average and standard deviation of the RMSE.
The L96 system is fully observed $\bH=\bI$ with a period of $\Delta t = 0.05$. The observations are independently perturbed with a normal distribution of error covariance matrix $\bR=\bI$.
The synthetic data assimilation runs make use of these observations only, over $5 \times 10^4$ time-steps, and after a burn-in of $10^4$ time-steps.
The average RMSE is computed by comparing the analysis of the DA runs with the true trajectory.
Finally, we do not account for a prior model error noise in these specific experiments ($\bQ =\bzero$).

In all experiments, the initial mean parameter vector is a perturbation of the true parameter vector (which is known in the L96 case) with an additive normal perturbation of standard deviation
$\sigma_a$. The value of $\sigma_a$ is taken in the range $[0, 0.64]$; $\sigma_a=0$ corresponds to the case where the model is known and hence to traditional DA with an EnKF.
The initial ensemble members are consistently generated by adding to this parameter vector estimate a random normal noise of standard deviation $\sigma_a$.
This perturbation is chosen to be very simple for the sake of clarity even though the parameters are quite different in nature.

\begin{figure}[t]
  \begin{center}
    \includegraphics[width=0.85\textwidth]{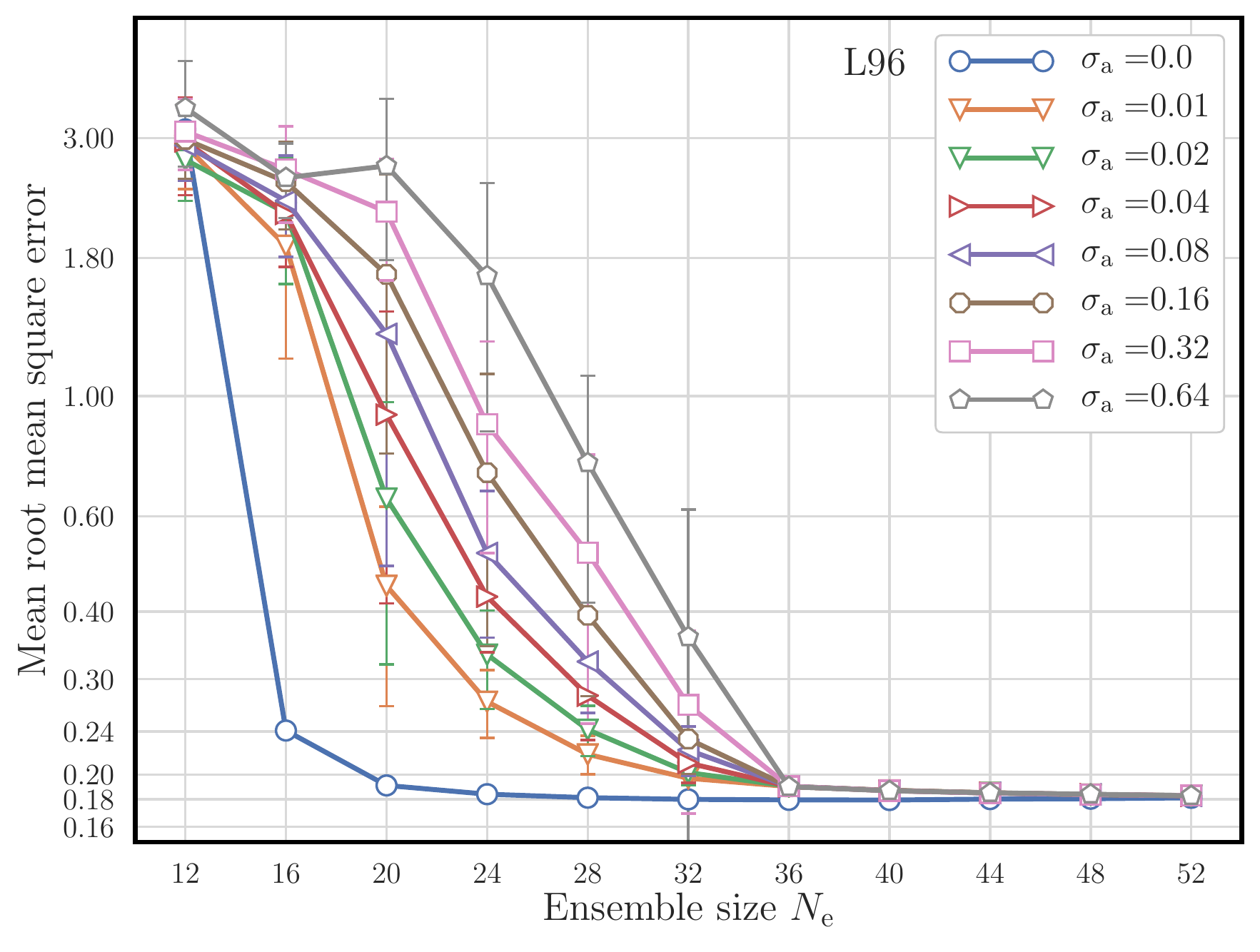}
  \end{center}
  \caption{\label{fig:1rev} Average state RMSE for the EnKF-ML with an hybrid adaptive inflation scheme, applied to the L96 model, for a range of ensemble size $\Ne$ and several values of the standard deviations of the initial parameter vector mean.  The error bars correspond to the standard deviation of $\Nexp=100$ repeated experiments.}
\end{figure}

The RMSE is plotted on Figure \ref{fig:1rev} as a function of the ensemble size, for several values of $\sigma_a$. A strong dependence of the RMSE on the initialization is observed whenever $\Ne \le \No+\Np$.
Quantitatively, we have $\No=14$ and $\Np=18$ for the L96 experiment, i.e., $\No+\Np=32$.
Moreover, due to its sensitivity to the parameter initial conditions, the RMSE has a high variability in this range, as shown by the error bars. When $\Ne > \No+\Np$, such an undesirable dependence vanishes.
This is accomplished with the help of the adaptive hybrid inflation scheme.
With higher values of $\sigma_a$, typically $\sigma_a \ge 1$, we may observe a divergence that would call for a fine tuning of the adaptive scheme.

This game also has limitations. If the prior information contained in the initial ensemble is insufficient, the DA run is bound to fail. One way to help is to use an approximate physical model as an initial guess for the model \cite{brajard2020b}. This might be required with higher dimensional and more complex models where a cold start (without any prior on the dynamics) would likely yield divergence.

\begin{figure}[t]
  \begin{center}
    \includegraphics[width=0.85\textwidth]{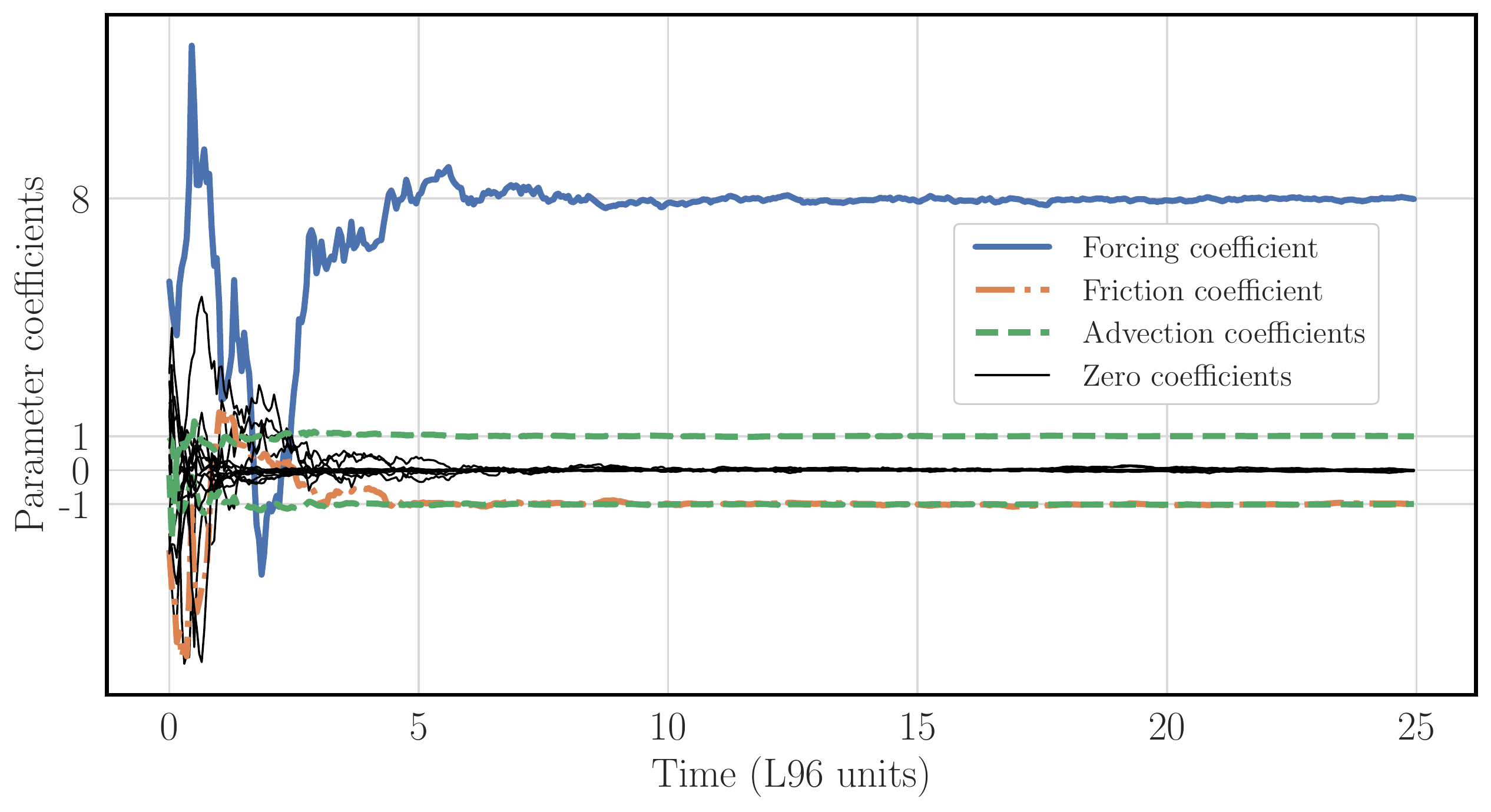}
  \end{center}
  \caption{\label{fig:2rev} Initial evolution of the $\Np=18$ parameters of the surrogate model learned on the observation of a L96 model run.
The key parameters are the forcing ($F=8$ in the true model), the friction ($-1$ in the true model) and the advection coefficients ($-1, 1$ in the true model).}
\end{figure}

A typical example of the evolution of the parameters in the burn-in stage is displayed in Figure \ref{fig:2rev} for the L96 model (where there is a one-to-one correspondence between the parameters of the true and surrogate dynamics). The setup is the same as above but with $\Ne=40$ and $\sigma_a=1$.

We also tested the impact of accounting for Gaussian model errors with prior error covariance matrix $\bQ=\sigma_q^2 \bI$ on the state vector,
using the deterministic SQRT-CORE algorithm \cite{raanes2015}. We chose ensemble sizes for which $\Ne \ge \No+\Np$ for both the L96 and L05III model, and we set $\sigma_a=0.2$ in both cases.
We observed that the stochastic noise does not really help, even in the less identifiable case of L05III.
It is likely that mitigating the impact of the surrogate model mismatch with the true dynamics is a role already played by the multiplicative inflation (adaptive or not). Additive inflation is overridden by multiplicative inflation in these experiments.
Hence, we will choose $\sigma_q=0$ for both the L96 and L05III cases hereafter. It must however be kept in mind that this conclusion might not apply to more complex models where multiplicative inflation does not necessarily optimally compensate for model error.

\subsubsection{Accuracy as a function of the ensemble size}

In the following, L96 and L05III are both considered. The setup for the L05III experiments is the same as that of the L96 experiments, with the same hyperparameters. However, for the L05III system, only the coarse variables are fully observed with a period of $\Delta t = 0.05$.

\begin{figure}[t]
  \begin{tabular}{cc}
    \includegraphics[width=0.47\textwidth]{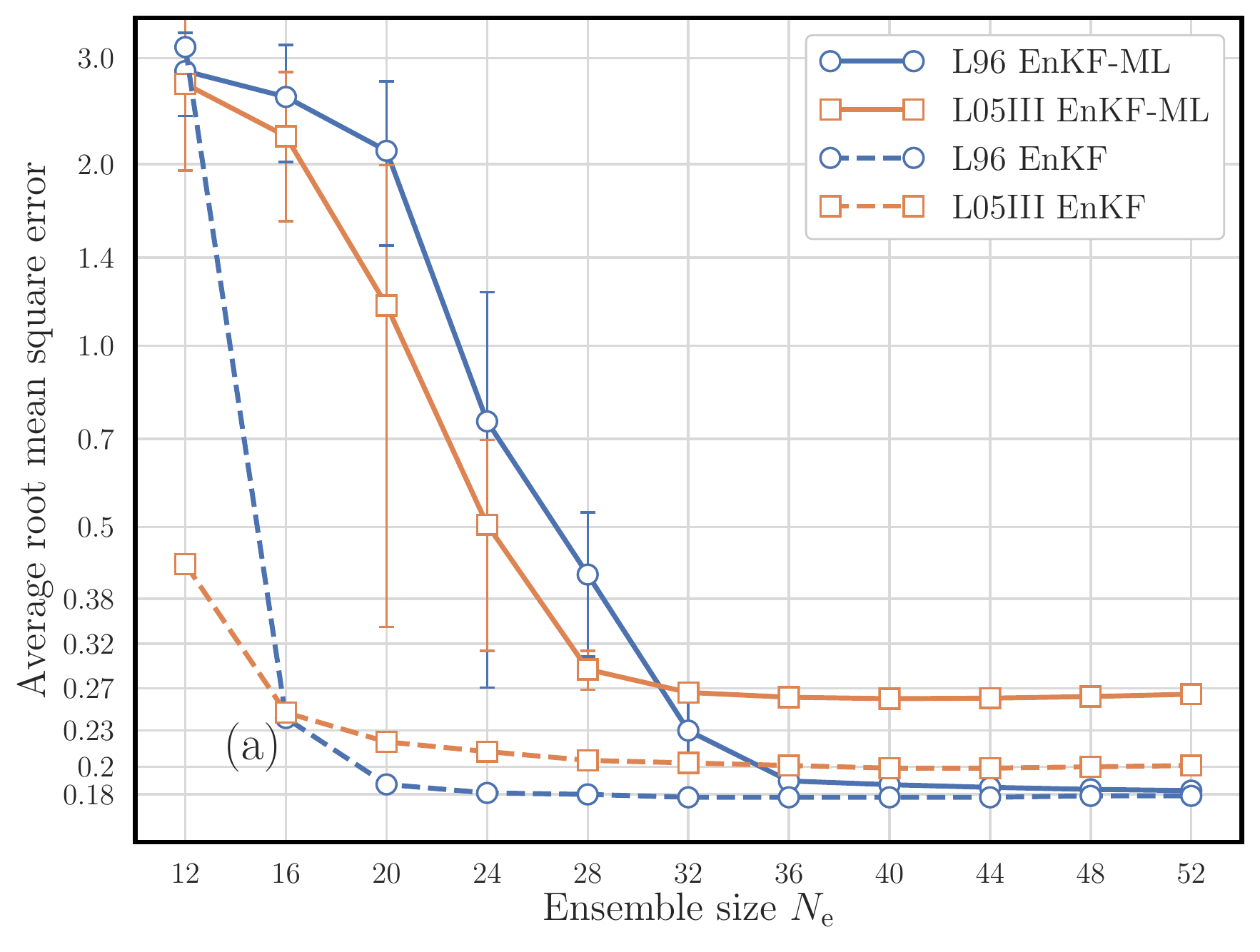} &
    \includegraphics[width=0.47\textwidth]{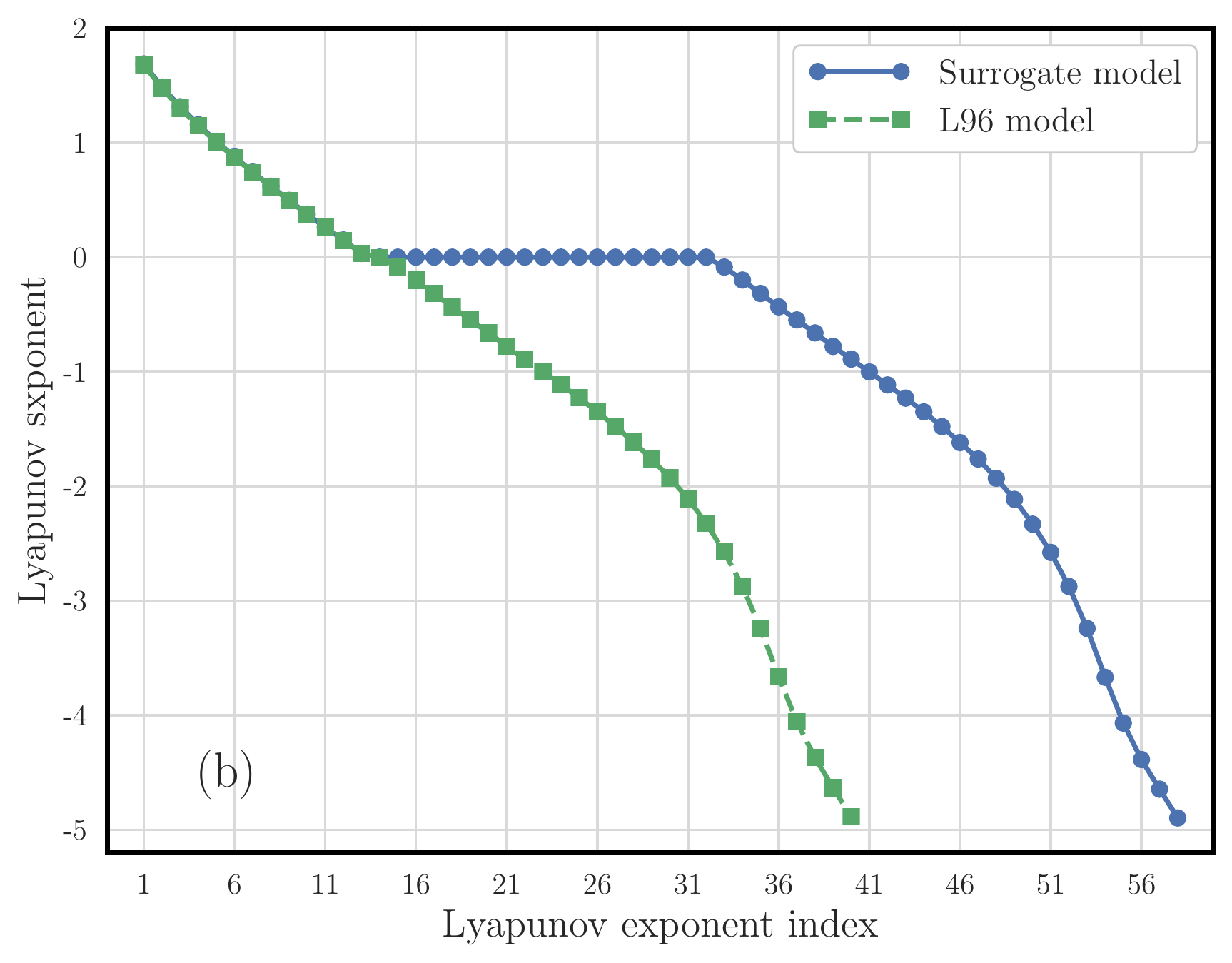}
  \end{tabular}
  \caption{\label{fig:3rev} Left panel (a): For both models, comparison of the performance of the EnKF-ML where the model is unknown and of the traditional EnKF where the model is known. The error bars are obtained from the standard deviation of $\Nexp=50$ repeated experiments.
 Right panel (b): Comparison of Lyapunov spectra of the L96 model and of the surrogate model about an L96 trajectory. }
\end{figure}

With these choices for the hyperparameters and adaptive inflation scheme, we obtain RMSE curves for the EnKF-ML and both models similar to the first experiments, using $\Nexp=50$ experiments. In addition, we compare the performance with the traditional ETKF where the model is known.
The results are shown in Figure \ref{fig:3rev}a.

In the L96 case, the performance achieved by the EnKF-ML is as good as that of the traditional EnKF in the critical region around $\Ne \ge 36$ (in the absence of localization) not far from the predicted $\Ne=33$ threshold. In the L05III case, the performance of the EnKF-ML is not as good as that of the EnKF. But considering that by the sole observations of the coarse modes, the true model cannot be identified, the accuracy achieved by the EnKF-ML is actually remarkable.

\subsubsection{Lyapunov exponents}
In order to numerically check the dynamical argument on the Lyapunov spectra which has consequences on localization (see Section~\ref{sec:enkf}), we compute the Lyapunov spectrum of the surrogate model about an L96 trajectory and compare it to the Lyapunov spectrum of the true L96 model. The spectra are plotted in Figure \ref{fig:3rev}b and corroborate our theoretical arguments. In particular, the positive part of the spectra coincides and the surrogate model spectrum has
as many neutral exponents as the number of dynamical parameters plus one because the model is autonomous.
Recall that we have $\No=14$ and $\Np=18$ for the L96 experiment, i.e., $\No+\Np=32$, and approximately the same for the L05III experiment. Note also that for $\Ne \ge \No+\Np+1 = 33$, the filters accuracy has reached its close to optimal RMSE value, as predicted in Sec.~\ref{sec:enkf}.

\subsection{The local EnKF-ML}
\label{sec:lenkf-ml}

In this section, we study the local EnKF-ML (LEnKF-ML). For the sake of limiting the number of experiments, we focus on the LEnSRF (local ensemble square root Kalman filter) implementation only, i.e. with CL, and leave testing the LETKF-ML based on DL for future work. We use a localization matrix based on the Gaspari--Cohn piecewise rational function \cite{gaspari1999}. The time interval between updates is $\Delta t = 0.05$. Unless otherwise stated, the synthetic observations are generated in the exact same way as in Section~\ref{sec:enkf-ml}, for both the L96 and L05III models. In particular, we do not account for prior model error ($\bQ =\bzero$). The hybrid adaptive inflation scheme employed in the global EnKF-ML is not used for the local EnKF-ML because it has not been tested on local EnKFs so far and would require further adaptation. Instead a scalar uniform multiplicative inflation is used. However we expect the regularization of localization to mitigate any residual dependency of the RMSE in the initial condition, as opposed to what was observed in the global EnKF-ML case.
The synthetic data assimilation experiments run over $10^4$ time-steps, after a burn-in of $5 \times 10^3$ time-steps, and are repeated $\Nexp=10$ times. Again, we choose $\sigma_a=0.2$. We first compute the mean RMSEs over the $\Nexp$ experiments with distinct random seeds
and later take the best RMSE over a range of multiplicative inflations, localization lengths and of state/parameter tapering coefficients. The uncertainty attached to any best RMSE is obtained from the $\Nexp$ values of the RMSE for the optimal values of inflation, localization length and tapering coefficient.

\subsubsection{Accuracy as a function of the ensemble size}

The accuracy of the LEnKF-ML is estimated via the RMSE, as a function of the ensemble size.
Keep in mind that the model parameters are not localized, so that a significant ensemble size should be required for convergence even with localization.
The LEnKF-ML is compared with the traditional LEnKF (in the shape of the LEnSRF here) based on the exact same setup but with the exception that the model is known. For the sake of numerical efficiency, the LEnSRF requires special attention in the L05III case,
where only the coarse modes are observed, i.e. $\Nx=36$ out of $\Nx+\Nu=396$ variables. The brute force implementation of the LEnSRF, \eqref{eq:leftT}, without augmented ensemble techniques \cite{bishop2017,farchi2019} becomes costly and we prefer to carry out the analysis expressed in observation space since (i) $\Ny/(\Nx+\Nu) \ll 1$ and (ii) the observations are local.
In that case, we can use (12) in \cite{bocquet2019b} which is the left-transform perturbation update but with algebra in observation space. As a consequence, we resort to the following convenient update formula for the state variables:
\begin{subequations}
  \begin{align}
    \bar{\bx}^\mathrm{a} &= \bar{\bx}^\mathrm{f} + \Bxy\T \left( \bR+\Byy \right)^{-1}\left( \by-\bH \bar{\bx  }^\mathrm{f} \right), \\
    \Xx^\mathrm{a} &=  \Xx^\mathrm{f} - \Bxy\left( \bR + \Byy
      + \bR\left(\bI + \bR^{-1}\Byy\right)^{\onehalf}\right)^{-1} \bY,
  \end{align}
  \end{subequations}
where $\Bxy = \Cxy \circ (\bX\bY\T)$ and $\Byy = \Cyy \circ (\bY\bY\T)$, using the fact that the observations are local so that the localization matrix blocks $\Cxy$ and $\Cyy$ can be straightforwardly defined.
The subsequent parameter update step is unchanged, i.e. \eqref{eq:ensemble-update-cl}.

\begin{figure}[t]
  \begin{tabular}{c}
    \includegraphics[width=0.85\textwidth]{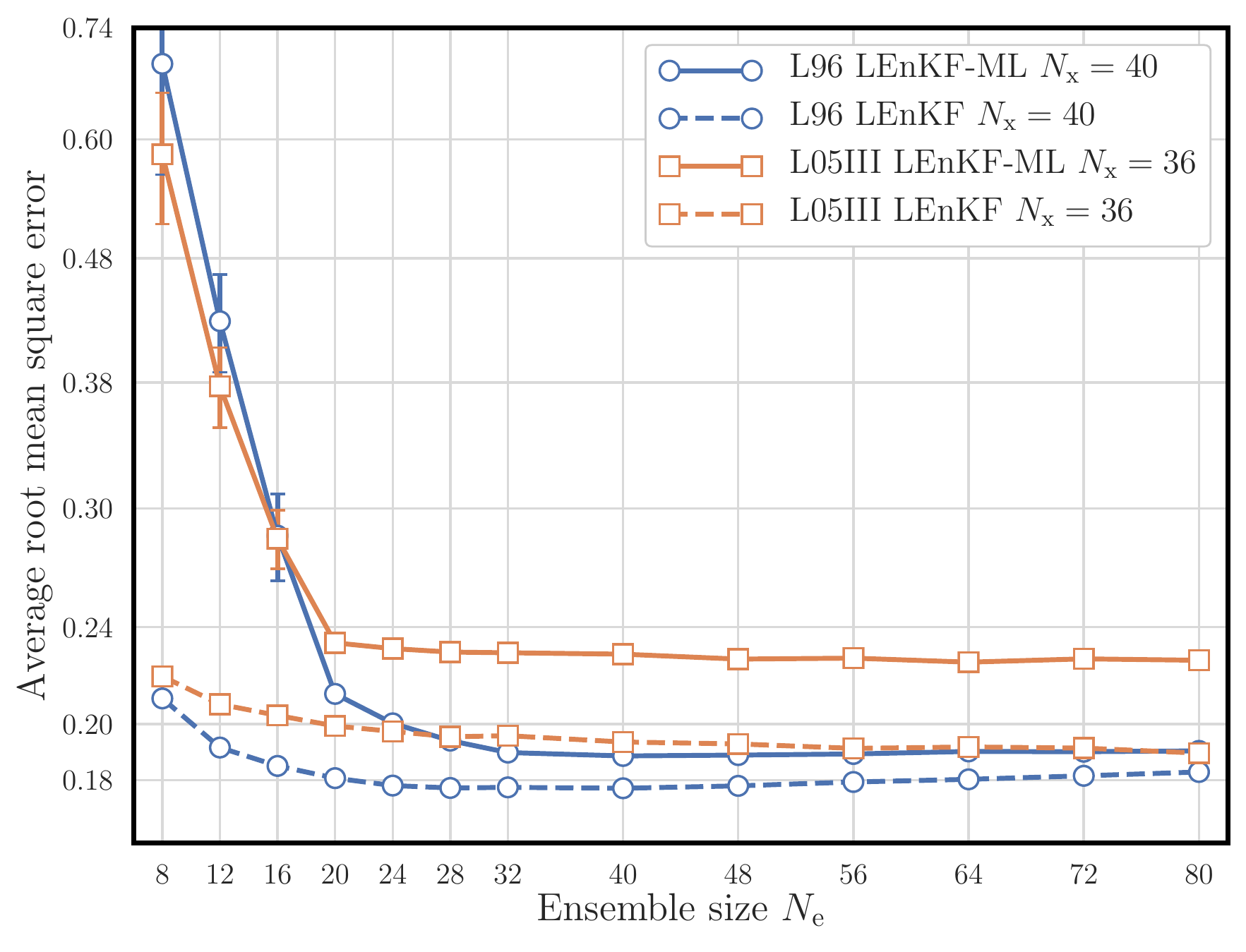}
  \end{tabular}
  \caption{\label{fig:4rev} For both models, comparison of the performance of the LEnKF-ML where the model is unknown and of the traditional LEnKF where the model is known.
 The error bars are obtained from the standard deviation of $\Nexp=10$ repeated experiments.}
\end{figure}

The results are shown in Figure \ref{fig:4rev}.
Since the error covariances of the parameters are not regularized, we can surmise that the LEnKF-ML should be efficient
for an ensemble size beyond $\Np$ plus a few additional members meant to deal with spatial state extension and on which localization applies, typically $5$ for both model cases. Hence the filter should reach good accuracy for $\Ne \ge \Np+5+1 = 24$, which is indeed what can be observed in Figure \ref{fig:4rev}. The accuracy in the L96 case almost reaches that of the traditional EnKF, the small RMSE gap being explained by the use of a uniform (rather than adaptive) inflation. The performance is slightly degraded compared to the traditional EnKF in the L05III case because of its non-identifiability when only the coarse modes are represented.

With the traditional LEnKF applied to the L96 model, an ensemble size of $5$ to $8$ with the above setup is sufficient to reach a good accuracy, while we need $\Ne \gtrapprox 24$ for the LEnKF-ML, which questions its scalability. To make sure that the LEnKF-ML is spatially scalable while keeping a good accuracy, we perform the experiment with the same setup as before but with an ensemble size of $\Ne=40$, a localization length of $18$ (defined as half-length of the Gaspari--Cohn function), a uniform multiplicative inflation of $1.005$ and the state space dimension of the L96 model is varied from $\Nx=80$ to $\Nx=1020$. Again, $\sigma_a=0.2$. Ensemble size, localization length and inflation are chosen close to optimal and fixed since we believe they are intensive quantities (as opposed to extensive).
As hoped for, the average RMSE remains within the interval $[0.186, 0.190]$ with a standard deviation of $3 \times 10^{-3}$ for $\Nx=80$ down to $5 \times 10^{-4}$ for $\Nx=1020$. We conclude that the LEnKF-ML is indeed scalable.

\subsubsection{Sensitivity to the observation density and observation noise magnitude}

In this section, we study the impact of the observation density $\Ny/\Nx$ on the accuracy of the LEnKF-ML, compared to the traditional LEnKF.
At each data assimilation cycle, $\Ny$ random grid points out of $\Nx$ (out of $\Nx$ coarse modes for L05III) are observed.
In a second experiment, we study the impact of the observation error standard deviation $\sigma_y$, assuming $\bR = \sigma^2_y \bI$, on the accuracy of the LEnKF-ML, compared to the traditional LEnKF, when all $\Nx$ grid points are observed (all the coarse modes in the L05III case). In both experiments, the ensemble size is set to $\Ne=24$.

\begin{figure}[t]
  \begin{tabular}{cc}
    \includegraphics[width=0.47\textwidth]{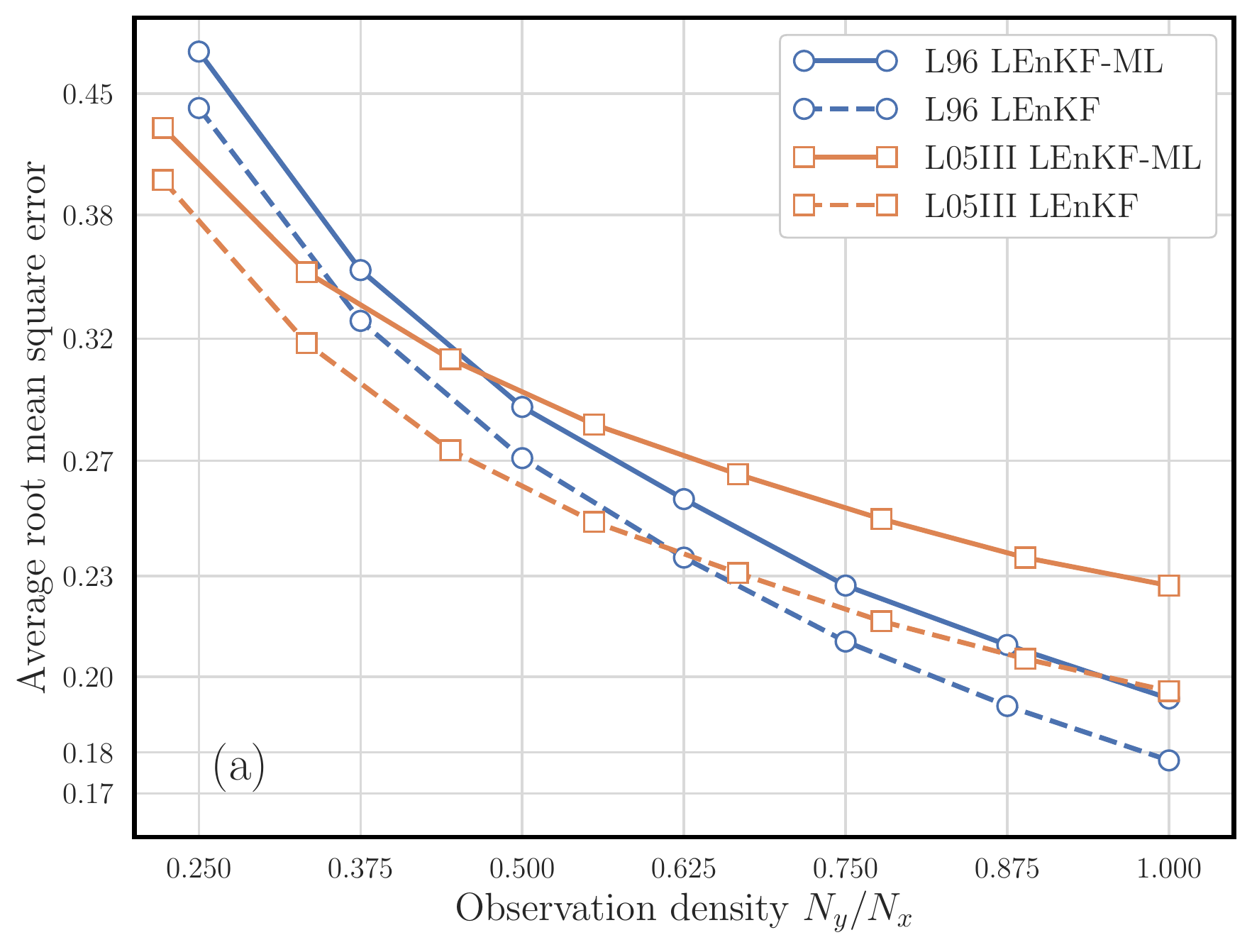} &
    \includegraphics[width=0.47\textwidth]{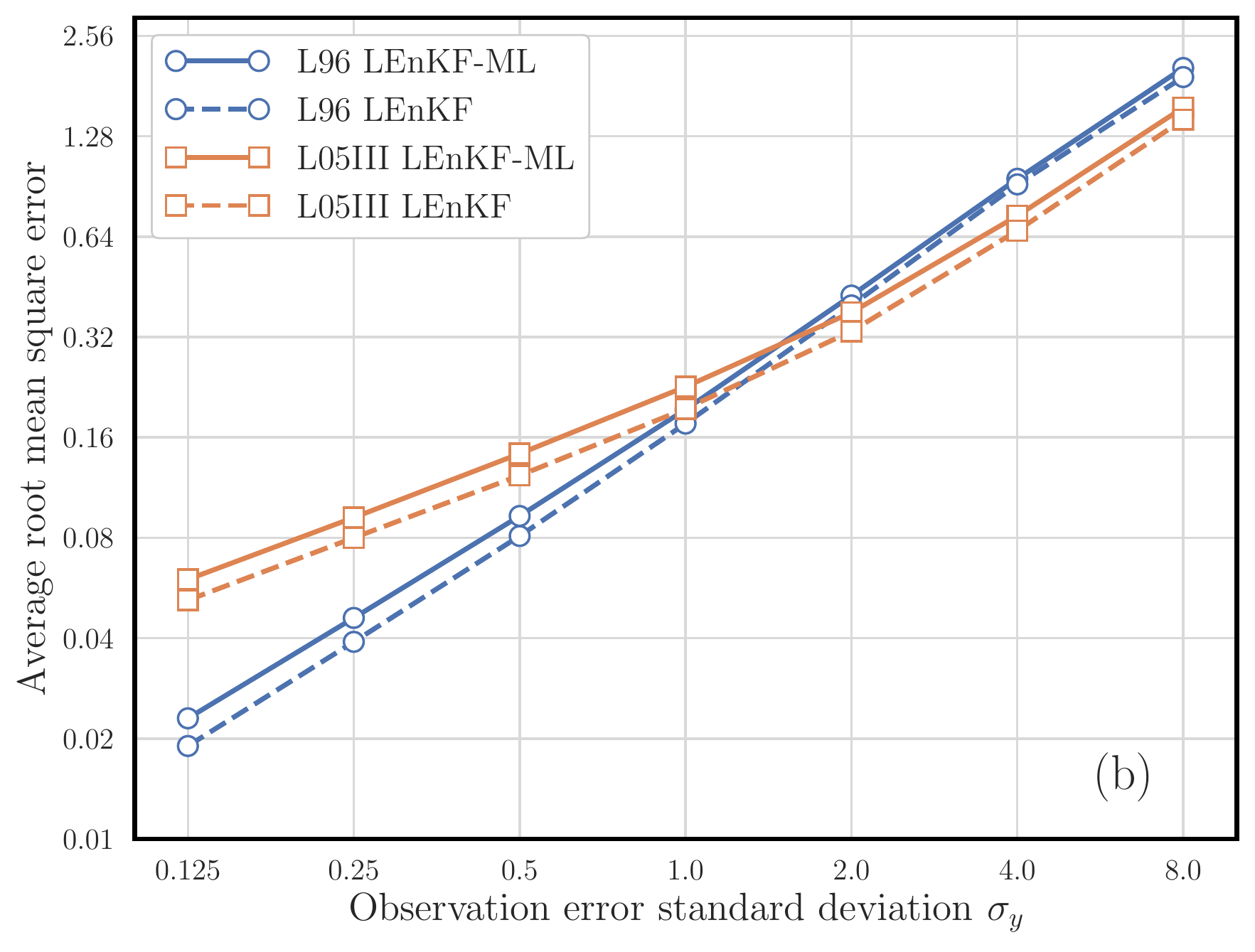}
  \end{tabular}
  \caption{\label{fig:5} Left panel (a): For both models, comparison of the performance of the LEnKF-ML where the model is unknown and of the traditional LEnKF where the model is known, as a function of the observation density.
    Right panel (b): For both models, comparison of the performance of the LEnKF-ML where the model is unknown and of the traditional LEnKF where the model is known, as a function of the observation error standard deviation. In all cases, the ensemble size is $\Ne=24$ and the RMSE statistics are accumulated over $\Nexp=10$ experiments.}
\end{figure}

Those are classical tests to evaluate DA techniques; they emphasize the potential of our approach in contrast to  standard ML techniques which are designed for fully and accurately observed systems.
The results are reported in Figure \ref{fig:5}. The figure displays the best average RMSEs over a range of localization lengths, tapering coefficients and multiplicative inflations.
They all show good performance of the LEnKF-ML filters for a wide range of hyper-parameters ($\Ny/\Nx$ and $\sigma_y$).
It is noticeable that, in relative terms (the y-axis of both panels are in logarithmic scale), the LEnKF-ML catches up with the traditional LEnKF reference
as the DA conditions become more dire (less observations, more noise).

\subsubsection{Optimal state/parameter tapering as a function of the ensemble size and state space dimension}

One intriguing problem is to determine the proper scaling of the optimal parameter/state tapering coefficient $\zeta$ with the ensemble size $\Ne$ and the state space dimension $\Nx$. Is the tapering of the cross state/parameter covariances a genuine regularization based on prior assumptions of these specific covariances, like traditional localization is? Is it just a mathematical requirement for the regularized covariance matrices to be semi-positive definite? Or is it an adjustment coefficient that tune the increment in parameter space as is clear from the update formula \eqref{eq:ensemble-update-dl-2} and briefly alluded to in \cite{koyama2010}?

\begin{figure}[t]
  \begin{tabular}{cc}
    \includegraphics[width=0.47\textwidth]{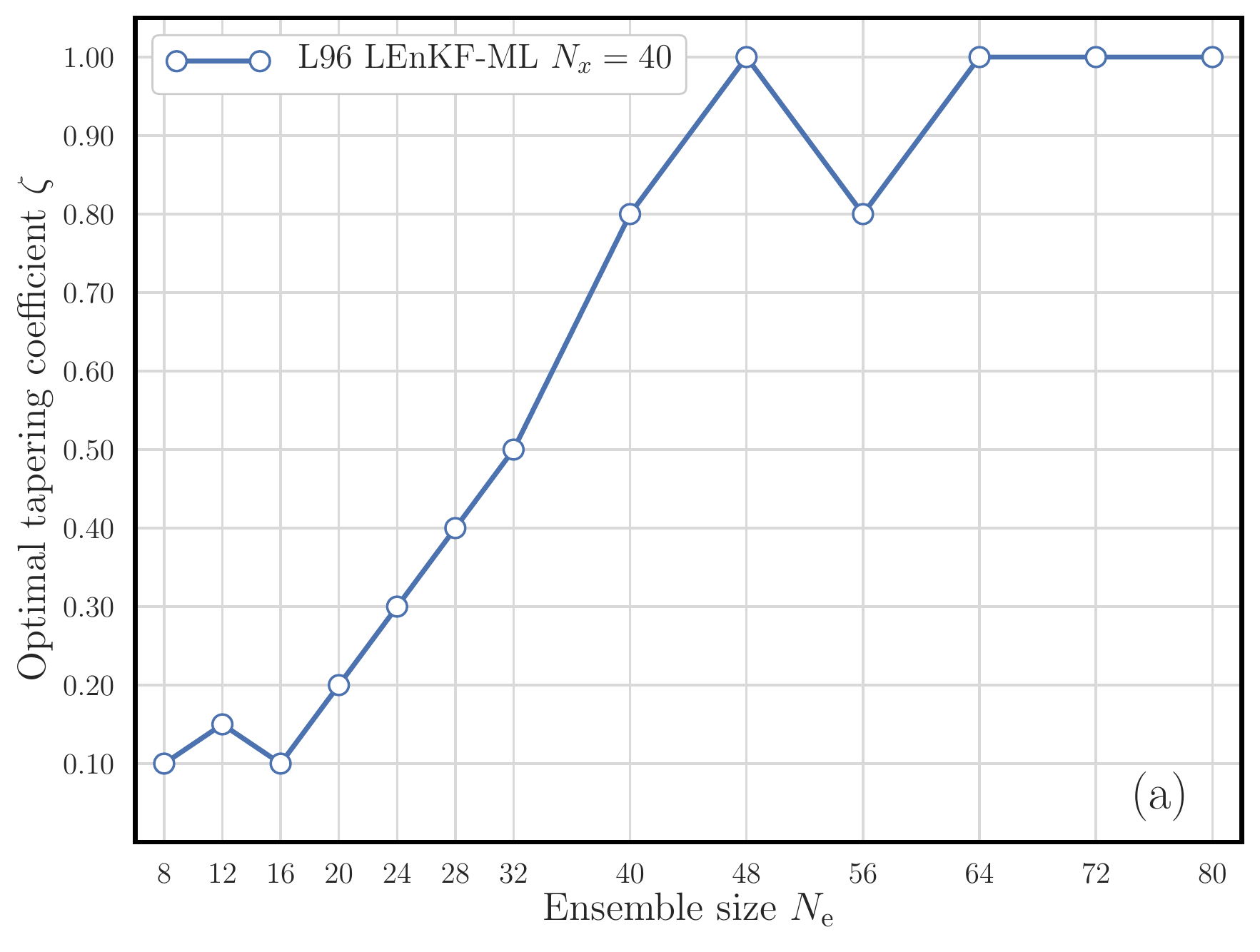} &
    \includegraphics[width=0.47\textwidth]{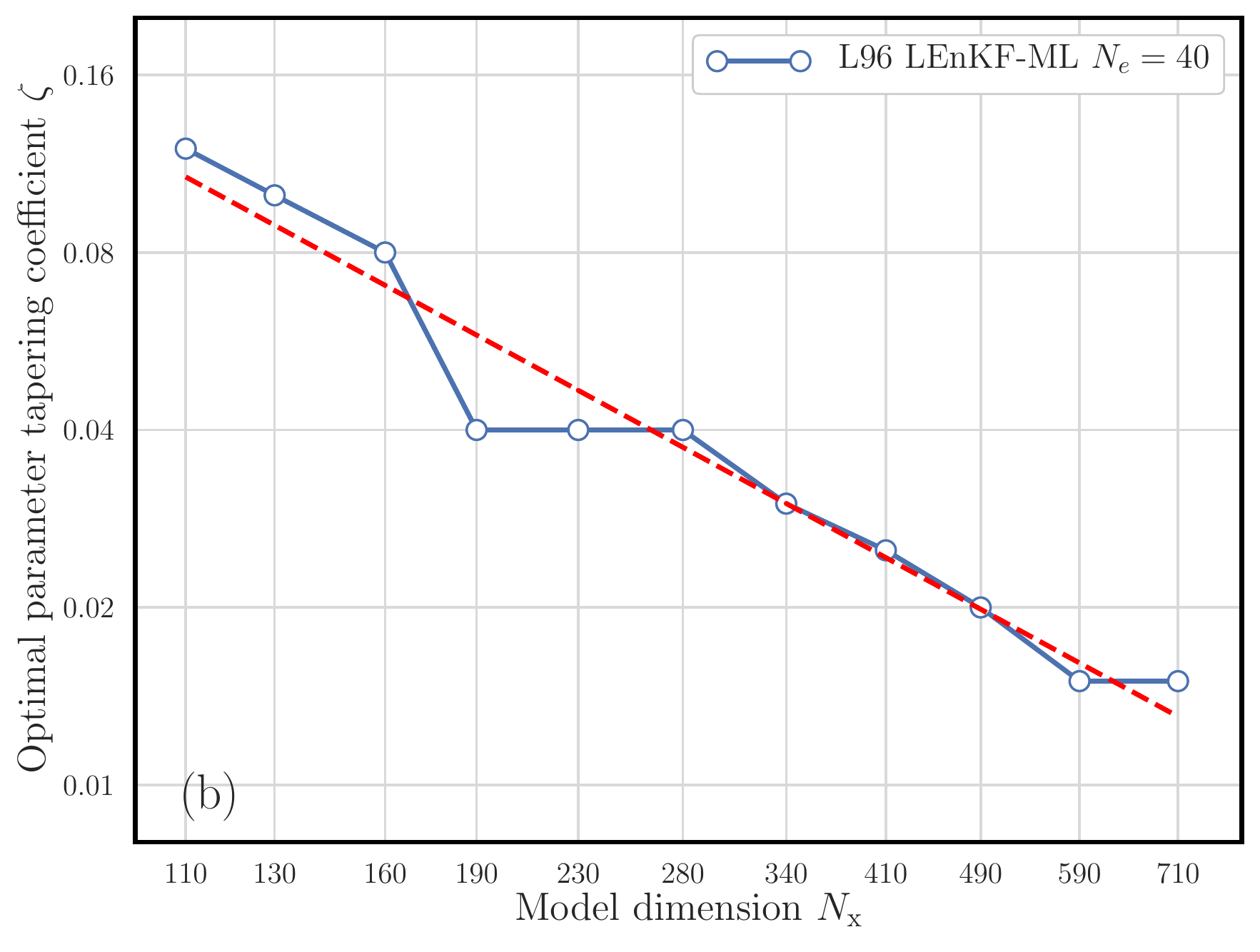}
  \end{tabular}
  \caption{\label{fig:6} Left panel (a): Optimal tapering coefficient $\zeta$ across a range of ensemble sizes $\Ne$ for LEnKF-ML applied to the L96 model ($\Nx=40$).
    Right panel (b): Optimal tapering coefficient $\zeta$ across a range of model state space dimensions $\Nx$ for LEnKF-ML applied to the L96 model, assuming $\Ne=40$, a fixed multiplicative inflation and localization length.} %% The error bars are obtained from the standard deviation of $\Nexp=10$ repeated experiments.
\end{figure}

The optimal tapering coefficient $\zeta$ is computed in the L96 model case only (for the sake of simplicity), using the LEnKF-ML.
Its dependence on the ensemble size at fixed model dimension $\Nx=40$ is studied.
Its dependence on the model dimension is then computed, while keeping the ensemble size fixed ($\Ne=40$),
using a fixed multiplicative inflation of $1.005$ and a localization length of $18$, since these optimal intensive hyperparameters
should not change significantly with the state space dimension. As before, we choose $\sigma_a=0.2$. The results are reported in Figure \ref{fig:6}.

At fixed state space dimension, the sampling errors in the full covariance matrices vanish, as the ensemble size increases.
Indeed, as can be seen in Figure \ref{fig:6}a, the optimal $\zeta$ increases towards $1$ and saturates at higher value ($\ge 0.70$) for $\Ne \ge 48$.
It was conjectured in \cite{ruckstuhl2018} and in our Section~\ref{sec:tapering} that the optimal $\zeta$ may asymptotically scale like the inverse of the dimension of state space. This is indeed close to what is observed in Figure \ref{fig:6}b. Asymptotically, i.e. for large enough state space dimension (here we have chosen $\Nx \ge 110$), the optimal $\zeta$ scales like $N^{-\alpha}$, where $\alpha=1.13 \pm 0.01$, as illustrated by the fit dashed red line in Figure \ref{fig:6}b.

\subsection{The iterative EnKF-ML}
For testing the iterative EnKF-ML, we do not consider localization since it would further add complexity. However, there is no fundamental obstacle to using localization
in a stronger nonlinearity context, although (i) it is more intricate than in the traditional EnKF context with a weaker nonlinearity (see \cite{bocquet2016,sakov2018}),
(ii) the techniques developed in Section~\ref{sec:lenkf} to consider global parameters and local variables should be implemented.
For evaluating the IEnKF-ML scheme, we use the same fully observed setup as before. The ensemble size is chosen to be $\Ne=40$ and $\Ne=36$ for the L96 and L05III cases, respectively.
The interval time between updates is varied uniformly from $\Delta t=0.05$ to $\Delta t=0.60$, which allows to increase the nonlinearity of the problem, in a regime where the iterative filters very significantly outperform the non-iterative filters. This is a classical setup for testing IEnKF methods \cite{sakov2012}.
However, as opposed to all the previous experiments, we consider the possibility of accounting for additive model error with a normal noise of error covariance matrix
$\bQ = \sigma_q^2 \bI$.
It is adequate to do so since the surrogate model is propagated over much larger time intervals, and might be subject to a lot of model error and biases. Considering  additive model noise is also part of the IEnKF-Q from which the IEnKF-ML is directly inspired. As control variable of the analysis step of the IEnKF-Q and now IEnKF-ML, this error is not implemented with SQRT-CORE which may have been employed in the (L)EnKF-ML.
However, as the model parameters get better and better estimated in the DA run, $\sigma_q$ can be made smaller and smaller.
For this reason, the optimal $\sigma_q$ (which minimizes the mean RMSE) is rather small in practice, and it is therefore here set to $0$, the multiplicative inflation compensating for residual model error anyway.
For the sake of simplicity, we do not use the hybrid adaptive inflation scheme so that there would be room for additional RMSE improvement.

First, we have observed that the performance of the IEnKF-ML for $\Delta t=0.05$ is not as sensitive to the initial ensemble of parameters as the EnKF-ML.
In particular, most IEnKF-ML runs are successful for $\sigma_a$ as large as $1$ (L96 case). Clearly, this is due to the smoother ability of the IEnKF-ML to estimate additive model error at each time step $\Delta t$.
However, unsurprisingly, this ability degrades rapidly as $\Delta t$ increases and nonlinearity becomes impactful.
Hence, for the sake of simplicity, we choose $\sigma_a=0.1$ in the following experiments.

\begin{figure}[t]
    \begin{center}
    \includegraphics[width=0.85\textwidth]{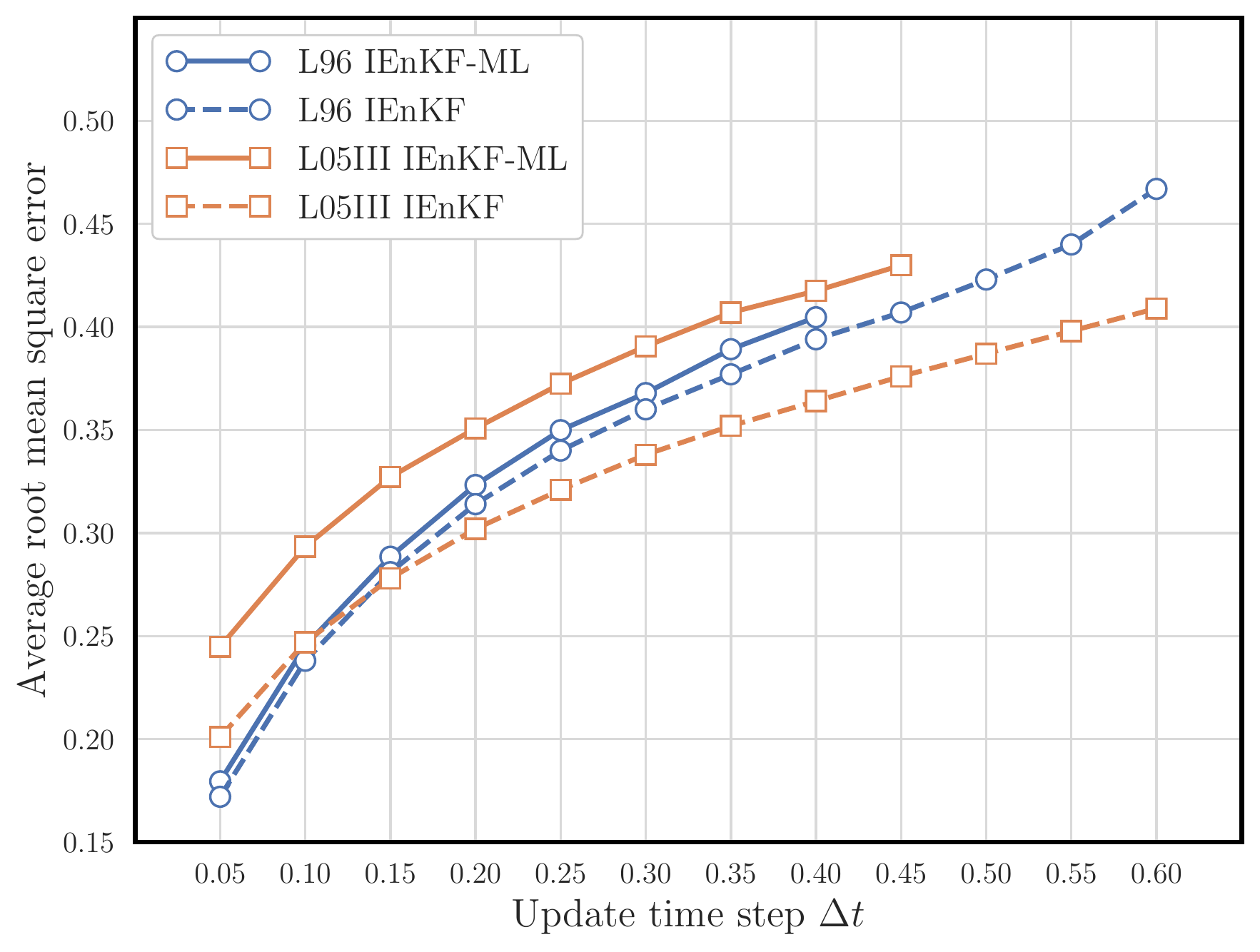}
  \end{center}
  \caption{\label{fig:7} Comparison for both models of the performance of the IEnKF-ML where the model is unknown and of the traditional IEnKF where the model is known, as a function of the time interval between updates $\Delta t$.
    The absence of a data point means that at least one of the $\Ne=10$ DA runs was divergent.}
\end{figure}

In Figure \ref{fig:7}, we report the best average RMSEs over the inflation, for both model cases.
These accuracy results are compared in each case to the traditional IEnKF where the model is perfectly known.
As expected, the gap of RMSE between the performance of the IEnKF-ML and the traditional IEnKF scheme in the L96 case is rather small while it is, as expected, larger for the L05III case.
Note that the IEnKF-ML yields RMSEs significantly lower than the EnKF-ML or even the traditional EnKF for large enough $\Delta t$ (not shown, see \cite{bocquet2012}, their Figure 3).
Moreover, they are stable up to large values of $\Delta t$, $0.40$ in the L96 case and $0.45$ in the L05III case.
Stability beyond these limits would require a smaller $\sigma_a$ (i.e. a more informative prior on the model) or an adaptive inflation scheme.

\section{Conclusion}
\label{sec:conclusion}

In this paper, we have studied the possibility to learn not only the state of a physical system from its partial and noisy observation, as in traditional DA, but also its full dynamics. This requires to use a parametric although quite general representation of the dynamics, such as a residual neural network implemented using a machine learning library.
We have extended the traditional EnKF, local EnKF and iterative EnKF to their machine learning counterparts where the model dynamics are learned. This is achieved sequentially, by successive updates as observation batches are collected.
These schemes are based on augmented state/parameter vectors and are hence strongly connected to state/parameter EnKF estimation.

The LEnKF-ML depends on the symmetries of the dynamics. When the dynamics are homogeneous, the model representation parameters are global, as opposed
to the state vector, and we developed an efficient two-step approach for updating the LEnKF-ML. Localization is not applied to the global parameters
but a tapering of the parameter/state variable cross covariances is necessary. On dynamical system grounds, it was explained that the ensemble size
should be large enough to represent the local degrees of freedom of the state vector (as in traditional DA) and the global model parameters.
In this case, the model representation should be minimal, typically a few dozens of parameters to efficiently represent homogeneous one or two-dimensional dynamics.

The IEnKF-ML is a surprisingly simple generalization of the traditional IEnKF, albeit in augmented state/parameter space.
Note, however, that the parameters are not only estimated through the cross-covariances with the observed state compartment, but also through the implicit nonlinear variational analysis of the IEnKF-ML. Because the sensitivities required by the analysis are generated by surrogate model propagations, the IEnKF-ML may turn much faster than the usual IEnKF if computationally efficient ML libraries are used.

A selection of these algorithms were numerically tested on the L96 model where the dynamics can theoretically be identified through an appropriate set of parameters, and the two-scale L05III model where only the coarse modes are observed and represented so that the true model cannot be accurately represented by the surrogate model. All tests were successful with good RMSEs typically $0-5\%$ and $0-15\%$ higher than with their traditional counterpart where the dynamics are exactly known, respectively. Using a hybrid adaptive multiplicative inflation scheme (hence accounting for both model and sampling errors), the RMSE can be almost as good as with a fully known model.

From this point on, there are many potential subjects of investigation.
Let us mention a few specific ones, among many.
First of all, we did not numerically evaluate the LETKF-ML, the DL counterpart to the LEnSRF-ML, although we describe the algorithmic parameter update step.
We plan on doing so.
Next, we would like to investigate the LEnKF with an inhomogeneous dynamics where not only the state but the model parameters are also local. Such LEnKF-ML is also scalable, and should not require much larger ensembles.
Third, in this paper, we have only focused on the filtering aspects of these online schemes. This is a limitation to a more efficient learning of the dynamics. To overcome this, we could consider sequential fixed-lag smoothers to better learn autonomous or slowly varying dynamics. Extending the EnKF-ML to classical smoother would yield the EnKS-ML. Extending the IEnKF-ML to longer DA time window would yield the IEnKS-ML. Note however that, as a lag-one smoother, the IEnKF-ML gave a glimpse of what could be achieved in this direction.
We do not anticipate further theoretical complications other than those mentioned in this paper.
We hope that this could significantly help in improving the accuracy of the surrogate dynamics.

%For acknowledgements section, please don't number the section, please begin it with \section*{Acknowledgements}
\section*{Acknowledgments}
The authors are grateful to two anonymous reviewers and Matthias Morzfeld, Editor, for their helpful comments and suggestions.
The authors acknowledge many insightful discussions on related topics with our colleagues Julien Brajard, Massimo Bonavita, Alberto Carrassi, Patrick Laloyaux, Laurent Bertino and Patrick Raanes. CEREA is a member of Institute Pierre-Simon Laplace (IPSL).

% We would like to thank you for \textbf{following
% the instructions above} very closely in advance. It will definitely
% save us lot of time and expedite the process of your paper's
% publication.

% You may incorporate your references as follows in your main tex file.
% Using BibTex is not recommended but can be handled.

%% \bibliographystyle{AIMS}
%% \bibliography{references.bib}

\providecommand{\href}[2]{#2}
\providecommand{\arxiv}[1]{\href{http://arxiv.org/abs/#1}{arXiv:#1}}
\providecommand{\url}[1]{\texttt{#1}}
\providecommand{\urlprefix}{URL }

\medskip
% The data information below will be filled by AIMS editorial staff
Received June 2020; revised August 2020.
\medskip

\end{document}